\documentclass[final,5p,times,twocolumn]{elsarticle}

\usepackage{mypreamble}
\usepackage{adjustbox}
\usepackage{tablefootnote}
\usepackage{textcomp, gensymb}

\begin{document}



\begin{frontmatter}

\title{Reduced Precision Floating-Point Optimization for \\ Deep  Neural Network On-Device Learning on MicroControllers}


\author[unibo,polito]{Davide Nadalini}
\ead{davide.nadalini@polito.it, d.nadalini@unibo.it}
\author[kuleuven]{Manuele Rusci}
\ead{manuele.rusci@esat.kuleuven.be}
\author[unibo,eth]{Luca Benini}
\ead{lbenini@iis.ee.ethz.ch}
\author[unibo]{Francesco Conti}
\ead{f.conti@unibo.it}


\address[unibo]{University of Bologna, Bologna, Italy}
\address[polito]{Politecnico di Torino, Italy}
\address[kuleuven]{Katholieke Universiteit Leuven, Leuven, Belgium}
\address[eth]{ETH Zurich, Zurich, Switzerland}

\begin{abstract}
{
Enabling On-Device Learning (ODL) for Ultra-Low-Power Micro-Controller Units (MCUs) is a key step for post-deployment adaptation and fine-tuning of Deep Neural Network (DNN) models in future TinyML applications. 
This paper tackles this challenge by introducing a novel reduced precision optimization technique for ODL primitives on MCU-class devices, leveraging the State-of-Art advancements in RISC-V RV32 
architectures with support for vectorized 16-bit floating-point (FP16) Single-Instruction Multiple-Data (SIMD) operations.
Our approach for the Forward and Backward steps of the Back-Propagation training algorithm is composed of specialized shape transform operators and Matrix Multiplication (MM) kernels, accelerated with parallelization and loop unrolling.
When evaluated on a single training step of a 2D Convolution layer, the SIMD-optimized FP16 primitives result up to 1.72$\times$ faster than the FP32 baseline on a RISC-V-based 8+1-core MCU. 
An average computing efficiency of 3.11 Multiply and Accumulate operations per clock cycle (MAC/clk) and 0.81 MAC/clk is measured for the end-to-end training tasks of a ResNet8 and a DS-CNN for Image Classification and Keyword Spotting, respectively -- requiring 17.1 ms and 6.4 ms on the target platform to compute a training step on a single sample. 
Overall, our approach results more than two orders of magnitude faster than existing ODL software frameworks for single-core MCUs and outperforms by 1.6 $\times$ previous FP32 parallel implementations on a Continual Learning setup.

}
\end{abstract}


\end{frontmatter}


\section{Introduction}
In recent years, the Internet-of-Things (IoT) ecosystem has been enriched by tiny battery-powered devices that can capture and locally analyze the sensed data~\cite{10.1145/3469029,wang2020deep}.
Notable examples include smart cameras for face recognition~\cite{Zemlyanikin_2019_ICCV}, nano-drones with autonomous navigation capabilities~\cite{lamberti_tinydronet}, hearable aids featuring noise cancelling~\cite{rusci2022accelerating}, wearable healthcare devices~\cite{8930945} or smart agriculture~\cite{9278587} systems, and more. 
To cope with the severely constrained energy budget and the form factor requirements, these \textit{smart} devices rely on MicroController Units (MCUs) as their main computational unit for processing the data coming from the tightly coupled sensors.
Differently from more capable engines such as edge GPUs or mobile-class CPUs (e.g., ARM Cortex-A multi-cores), 
MCUs present a power envelope lower than a few hundred mW to comply with battery-powered operation. 
On the other side, running complex processing pipelines, i.e., based on modern Deep Neural Networks (DNNs), on these platforms can be extremely challenging because of the limited compute and memory budget, which typically amounts to only a few MBs of on-chip memory~\cite{banbury2020benchmarking}. 

The commonly adopted design flow to bring DNN inference models to low-power MCUs is composed of an initial training phase, typically performed in a GPU-equipped data-center machine, followed by the deployment of the frozen trained model on the end-point device.
This rigid scheme, indicated as \textit{train-once-deploy-everywhere}, has started to be questioned because of the lack of robustness observed when testing smart devices in the real world. 
A major error source concerns the nature of the data sensed in the field that differs substantially from the training data, e.g.,  when a device is sensing an unknown environment not well-represented in the train set~\cite{https://doi.org/10.48550/arxiv.1606.06565}.
Because of this mismatch, the prediction accuracy can be drastically reduced compared to the accuracy scored on the test dataset used at design time.  
Transfer Learning~\cite{weiss2016survey} or recently proposed Continual Learning~\cite{de2021continual} techniques address this issue by fine-tuning the trained DNN, i.e., updating the model coefficients, over new data coming from a new domain. 
Unfortunately, these adaptive solutions cannot scale if relying on external servers for the training tasks, considering that every individual device can face a different domain that may be subjected to rapid changes. 

To address this challenge, we focus on the \textit{On-Device Learning} (ODL) paradigm~\cite{konevcny2016federated}.
According to this, end-point devices rely on the local compute capacity for the (incremental) learning task rather than running inference-only workloads.
Instead of continuously exchanging data and parameters between nodes and a central server for the DNN model updates, the ODL policy reduces communication costs and lowers the workload on the server side.
Additionally, ODL brings benefits to bypass major privacy concerns: local execution prevents the sharing of personal (labeled) data to third-parties cloud services used for the training process.

All these points motivate us to address the fundamental research question concerning the \textit{feasibility of On-Device Learning on low-end IoT devices} powered by tiny edge devices such as MCUs.
We refer to a learning scheme using the Back-Progagation (BP) algorithm, a gradient-based optimization strategy typically used for DNN training. Several recent works targeting resource constrained devices addressed this problem by introducing hard restrictions to the BP algorithm.
%
The TinyOL~framework\cite{9533927} and the STM32 NanoEdge AI Studio\footnote{STM32 NanoEdgeAI: https://www.st.com/en/development-tools/nanoedgeaistudio.html}, for example, enable learning on MCU by updating the parameters - i.e. fine-tuning - either of only the last layer of the deployed DNN, or in limited scenarios, with respect to in-field data. Here, the BP computation leverages full-precision floating-point (FP32) arithmetic.
The rest of the model, previously quantized to low-bitwidth, is kept frozen.
Restricting training to the last layer drastically reduces the expressiveness of the method, i.e., the complexity of what can be learned, limiting the overall effectiveness.
On the other hand, Tiny Training Engine~\cite{lin2022device} applies gradient scaling to use 8-bit arithmetic for the backward pass in combination with a sparse weight update logic. 
This approach covers the entire network but still compromises between the training complexity and its efficacy; the algorithm cannot have, therefore, the same general applicability as conventional BP.
Only a recent work, \textit{AIfES}\footnote{AIfES for Arduino: https://github.com/Fraunhofer-IMS/AIfES\_for\_Arduino}, has focused, instead, on deploying the full BP algorithm on MCU. 
This library covers several full-precision operators, emphasizing completeness at the expense of speed, as it does not support any optimization for multi-core execution, optimal loop unrolling, and half-precision floating-point execution.

In this work, we explore the feasibility of Back-Propagation-based ODL on an ultra-low-power device from a novel perspective, providing an in-depth exploration of software-based optimization strategies to accelerate the full BP algorithm for multi-core MCUs at the frontier of the State-of-the-Art.
First, we propose a comprehensive computational analysis of the basic primitives required for training a DNN on an MCU, decomposing them in shape transform operations (e.g., \textit{Im2Col}) combined with Matrix Multiplications (MM) -- essentially, extending to training the work conducted by Lai~et~al.~\cite{lai2018cmsisnn} for DNN inference.
Second, leveraging recent advances in MCU architecture design, we deploy our work on a 22nm silicon embodiment of the Parallel Ultra Low Power (PULP) platform~\cite{rossi2021vega}, GreenWaves GAP-9, which is a RISC-V multi-core design with support for SIMD-accelerated half-precision (16-bit) floating-point (\texttt{FP16}).
We ascertain whether the architectural improvements related to parallelism, reduced precision, and SIMD translate to proportional improvements in performance and energy efficiency, compared both to a single-core highly optimized full-precision floating-point (\texttt{FP32}) baseline tested on the same platform and on a single commercial STM32 MCU.
In particular, we design a set of optimized software primitives leveraging FP16 arithmetic, which is nowadays widely adopted on the server side for efficiently training DNN models without accuracy penalties with respect to FP32 \cite{cheng2019distributed,kalamkar2019study,micikevicius2018mixed}. Furthermore, the choice of the specific FP16 format can be tuned by the user in accordance with the target device's specifications (e.g., Vega~\cite{rossi2021vega}, which supports both IEEE FP16 and Bfloat16).
Finally, to investigate whether the proposed optimization can make ODL feasible in realistic use cases, we consider the class-incremental Continual Learning  case study proposed by Pellegrini~et~al.~\cite{pellegrini2020latent}, and we compare, in terms of latency and energy consumption, the solutions obtained using AIfES -- the most complete MCU training framework currently available -- and the proposed library.

In detail, this work makes the following contributions towards the State-of-the-Art for MCU-based ODL:
\begin{itemize}
    \item We analyze the training primitives for a DNN, focusing on the Conv2D case, and derive foundational abstractions for the basic  operators, discussing the impact of data layout (channel-height-width / CHW vs. height-width-channel / HWC) on the underlying computational structure.
    \item We introduce latency-optimized software primitives for MM kernels exploiting loop unrolling, parallelization, and SIMD FP16 and introducing transposed MM (MM$_T$) and \textit{Im2Row} transformations to minimize transposition overheads in SIMD-vectorized MM.
    \item We analyze in detail the latency impact of transform operators needed by every training kernel and quantify the impact, in terms of latency, on the learning task.
    \item We assess the execution latency and the energy consumption of our primitives on individual DNN layers, comparing baseline and optimized layers on the target GreenWaves GAP9.
    \item We explore optimized MM primitives, inspired by the same principles, on an STMicroelectronics STM32L4 to provide a further testing point for our approach.
    \item We compare the training of the end-to-end case study proposed by Pellegrini~et~al.~\cite{pellegrini2020latent} between our proposed framework on GAP9 and \textit{AIfES} on STM32L4.
\end{itemize}

Our optimized MM functions, which are the core kernels of the proposed ODL primitives, achieve a peak performance of 7.89 MAC/clk on GAP9 when leveraging FP16 SIMD instruction and 8-core parallelism, 1.91$\times$ faster than the FP32 counterpart. 
When benchmarking a complete training step of a Conv2D layer, the computational efficiency reduces to 6.62 MAC/clk because of the overhead of the shape transform functions, which impact 12.5\% of the computational time. 
Such overhead is mitigated by using an HWC data layout, which is 11\% faster than a CHW-based implementation. 
Overall, a latency of 17.1 ms and 6.4 ms is accounted to run the forward and backward steps for a ResNet8 for Image Classification, and a DS-CNN for Keyword Spotting on a GAP9 SoC clocked at 370MHz while consuming 60.5 mW on average. 
Our evaluation of a Continual Learning case study shows that our solution results 1.63$\times$ and 767$\times$ faster than previous solutions based on, respectively, FP32 training primitives running on the same platform and a single-core MCU using the open-source \textit{AIfES} library.

To foster future research on MCU-based On-Device Learning, we release the code of our library as open-source software at: \url{https://github.com/pulp-platform/pulp-trainlib}.


\section{Related Work}

\begin{table*}[t]
\caption{On-Device Learning Methods for Ultra-Low-Power Devices}
\resizebox{\textwidth}{!}{%
\begin{tabular}{|p{3cm}| p{4cm}|p{3cm}|p{5cm}|p{4cm}|p{2cm}|}
\hline
\textbf{ODL Method}  & \textbf{Target Task}                                                              & \textbf{Retrainable Layers} & \textbf{Variations to Backpropagation}                              & \textbf{Target Device}          & \textbf{Data Type} \\ \hline
De Vita \cite{9903209}  & Anomaly Detection                                                                 &  ESN layer                   & Custom Echo State Network (ESN) for time series                     & STM32 boards                    & FP32               \\ \hline
TinyOL \cite{9533927}  & Image Classification                                                              &   Only last layer             & Extra custom trainable layer on bottom of a frozen DNN              & Arduino Nano 33 BLE             & FP32               \\ \hline
TinyTL \cite{https://doi.org/10.48550/arxiv.2007.11622} & Image Classification                              &  All (biases only)           & Bias training only (reduction in activation size)                   & Generic embedded device         & FP32               \\ \hline
Ravaglia \cite{9580920}   & Image Classification                                                              &   Last N layers               & Continual Learning with quantized Latent Replays                    & Vega \cite{rossi2021vega}, STM32L4                   & FP32               \\ \hline
Train++ \cite{9604425} & Binary Classification                                                             &   All                         & Custom incremental learning algorithm for binary classification     & ARM Cortex-equipped MCUs, ESP32 & FP32               \\ \hline
PocketNN \cite{https://doi.org/10.48550/arxiv.2201.02863} & Image Classification                                                              &  All                         & Integer-Only Direct Feedback Alignment (no Backprop)                & Generic edge device             & INT8               \\ \hline
Tiny Training Engine \cite{https://doi.org/10.48550/arxiv.2206.15472} & \begin{tabular}[c]{@{}l@{}}Image Classification / \\ General Purpose\end{tabular} &   All (quantized)             & Automatic gradient scaling to fit INT8 precision + gradient pruning & STM32F746, Other MCUs           & INT8               \\ \hline
\end{tabular}
}
\label{tab.ODLMethods}
\end{table*}

\begin{table*}[t]
\caption{Backpropagation-Based On-Device Learning Implementations on MCUs}
\resizebox{\textwidth}{!}{%
\begin{tabular}{|p{4cm}| p{4cm}|p{3cm}|p{4cm}|p{4cm}|p{2cm}|}
\hline
\textbf{ODL Implementation} & \textbf{Target Task} & \textbf{Retrainable Layers} & \textbf{Kernel Optimizations}               & \textbf{Target Device}                         & \textbf{Data Type}  \\ \hline
ODDA  \cite{9869990}  & Keyword Spotting     & All                         & None                                        & Vega \cite{rossi2021vega}, Raspberry PI-4B, Snapdragon 888          & FP32                \\ \hline
Giménez \cite{9797171,electronics11040573} & Keyword Spotting     & All (Fully-Connected)       & None                                        & Arduino Nano 33 BLE, Arduino Portenta H7       & FP32               \\ \hline
AIfES \cite{AIfES}  & General Purpose      & All                         & Matrix Multiplication (ARM CMSIS-NN)        & Arduino boards, ARM Cortex-M cores             & FP32               \\ \hline
PULP-TrainLib \cite{10.1007/978-3-031-15074-6_13} & General Purpose      & All                         & Matrix Multiplication (FP32)                & RISC-V Multicore MCUs, STM32 boards            & FP32             \\ \hline
\textbf{This Work}  & General Purpose      & All                         & Matrix Multiplication, Im2Col/Im2Row (FP16) & RISC-V Multicore MCUs, MCUs with FP16 SIMD FPU & FP16              \\
\hline
\end{tabular}
}
\label{tab.ODLImplementations}
\end{table*}

\subsection{On-Device Learning on Tiny Edge Devices}
To review the existing techniques, we first analyze  lightweight ad-hoc methods for ODL. Secondly, we describe the existing ODL applications and implementations targeting MCU devices.

\subsubsection{Restrictions to Backpropagation}

Several works, which we summarized in Table~\ref{tab.ODLMethods}, address the ODL problem by focusing on the reduction of the computational burden of the Backpropagation (BP) algorithm, by applying specific restrictions or directly replacing BP with a proxy. 
%
Focusing on time series analysis for Anomaly Detection, De Vita~et~al.~\cite{9903209} extended the functionalities of STMicroelectronic's X-CUBE-AI\footnote{STM X-CUBE-AI: https://www.st.com/en/embedded-software/x-cube-ai.html} by introducing support for On-Device Training of Echo State Networks. The method was tested on an STM32 MCU featuring less than 100 kB of memory occupation. 
%
To enable lightweight transfer learning, TinyOL \cite{9533927} proposes to insert a single trainable layer on top of a frozen and quantized model. 
This extra layer is trained in a few milliseconds using ARM-Cortex-equipped Arduino boards in both supervised and unsupervised setups. 
%
Similarly, Train++ \cite{9604425} implemented ODL for on-device targets but targeted shallow single-layer networks for binary classification problems.
%
To reduce the memory footprint of the activations tensors for the training task, TinyTL \cite{https://doi.org/10.48550/arxiv.2007.11622} proposes to limit the backpropagation to biases only.
Within a transfer learning context, this approach can reduce the memory requirements by up to 12.9$\times$ with respect to training also the weight parameters at the cost of an extra custom residual layer for preserving the accuracy level. 
These works only train a subset of the weight parameters to prevent the implementation of costly Backpropagation algorithms on resource-constrained MCUs. 
In contrast, we address this challenge by developing an optimized software methodology that exploits advanced multi-core MCU designs with reduced-precision FPU support.


To bring ODL to resource-constrained devices lacking FPU support,  PocketNN \cite{https://doi.org/10.48550/arxiv.2201.02863} presented a training methodology to exploit integer-only computation based on Direct Feedback Alignment \cite{NIPS2016_d490d7b4}.
%
To reach the same goal, Tiny Training Engine (TTE) \cite{https://doi.org/10.48550/arxiv.2206.15472} combined gradient tensors pruning via offline calibration and a novel Quantization-Aware strategy for scaling the gradient magnitude and fitting the limited integer range. 
Thanks to this approach, the authors demonstrated a training procedure for low-end MCUs leveraging 8-bit computation kernels. 
In contrast to these approaches, our work does not impose modifications or custom training algorithms for the learning process for ODL, nor does it require an additional offline calibration procedure. Rather, we support and accelerate the canonical and commonly used Backpropagation to broaden the scope of ODL without paying accuracy degradations due to limited integer ranges \cite{gholami2021survey}. 

\subsubsection{ODL Implementations for MCUs}

Table~\ref{tab.ODLImplementations} reports the works addressing the application and implementations of complete Backpropagation on Ultra-Low-Power MCUs.
Targeting the problem of noise domain shift in audio keyword spotting, Cioflan~et~al.~\cite{9869990} propose to increase the accuracy of their classification model using On-Device Domain Adaptation (ODDA). 
Thanks to this strategy, they could achieve an accuracy improvement by 1.43\% at a memory cost of only 1.47 MB; still, the latency higher than 100s prevented real-time application.
%
Giménez~et~al.~\cite{9797171} used simple DNNs composed of Fully-Connected layers to learn in-the-field simple audio commands  in several milliseconds with tiny MCUs. 
The same authors \cite{electronics11040573} later extended this approach to a distributed setup using Federated Learning \cite{LI2020106854,9060868}.
This method presents a memory footprint of less than 256 kB but requires a training latency of several hundreds of seconds for a full federated update.
These works focused on the applications of ODL rather than addressing performance optimization, as we consider in our work. 

In contrast, a small group of works targeted the design of a complete training framework for ODL on MCUs. 
AIfES by Fraunhofer IMS is currently a state-of-the-art library for ODL on Arduino and ARM Cortex-based MCUs, supporting   Fully-Connected and Convolutional layers and a variety of  activation functions, optimizers, and commodity functions for training. 
They rely on CMSIS-DSP MM kernels for latency-optimized ODL. 
%
PULP-TrainLib \cite{10.1007/978-3-031-15074-6_13} is an ODL framework for RISC-V Multicore MCUs, featuring a set of FP32 performance-tunable training primitives of Fully-Connected and Convolutional layers. To find the fastest configuration for each training step and DNN layer, PULP-TrainLib employs an Autotuner to select the fastest MM algorithm.
Ravaglia~et~al.~\cite{9580920} exploited an early prototype of the PULP-TrainLib to demonstrate Continual Learning (CL) for image recognition on MCUs. 
In this work, we leverage the PULP-TrainLib templates and extend them with novel latency-optimized software primitives that take advantage of multi-core RISC-V MCUs with FP16 SIMD support. To the best of our knowledge, our design results in the fastest ODL library for MCU targets.

\subsection{HW Support for Reduced Precision}
The opportunity to accelerate the computation by exploiting low-bitwidth precisions is fostering the research community, pushing for new HW concepts and MCU sub-systems.
In this context, J. Lee~et~al.~\cite{8662302} presented LNPU, a Sparse DNN Processor which allows Fine-Grained Mixed Precision between FP16 and FP8 to enable on-chip training. LNPU features 16 sparse Deep Learning cores, orchestrated by a single Central Core, a SIMD core, and a RISC controller, with a power consumption of 43.1 to 367 mW, at an operational frequency of 50 and 200 MHz, respectively. Furthermore, LNPU features a peak efficiency of up to 25.3 TFLOPS/W, while processing inputs with 90\% of sparsity.
Targeting RISC-V cores as the main computational cores, F. Montagna~et~al.~\cite{9506919} present a multi-core transprecision computing cluster that aims at minimizing the power consumption of near-sensor applications. The authors combine hardware sub-word vectorization and a dedicated interconnect to efficiently share multiple Floating Point Units (FPUs) among up to 16 parallel cores while providing a complete software infrastructure to enable efficient parallel programming. In terms of performance, their solution achieves a peak performance of 2.9 GFLOPS with a power consumption of 43 mW, which is compatible with the low-power environment.
Other works focus on general-purpose strategies to provide Reduced Precision on Ultra-Low-Power SoCs.
Among the RISC-V\footnote{RISC-V International: https://riscv.org/} MCU architectures, D. Rossi~et~al.~\cite{rossi2021vega} presented Vega, a ten-core System-on-Chip (SoC) based on the Parallel Ultra-Low Power (PULP) platform\footnote{PULP Platform: https://pulp-platform.org/}. Vega's cores are equipped with a set of floating-point units (FPU) capable of dealing with different floating-point formats, as wide as 32-bit and Single-Instruction Multiple-Data (SIMD) 16-bit, as well as two programmable Machine Learning accelerators. Thanks to these features, Vega achieves a State-of-the-Art performance of up to 
129 GFLOPS/W for FP16 computations. In this paper, we refer to this HW concept to design a set of software primitives optimized to exploit the Reduced Precision FPUs for DNN training tasks. On the other side, 
the recently introduced ARMv8.1-M\footnote{ARMv8 M-Profile Architecture: https://developer.arm.com/Architectures/M-Profile\%20Architecture} enabled FP16 processing as part of the new Helium M-Profile Vector Extension (MVE), which allows up to $15 \times$ speedup on Machine Learning applications and $5 \times$ speedup on DSP, with respect to ARMv8 instructions. However, to the best of our knowledge, no off-the-shelf MCU equipped with ARM MVE is yet available on the market.
%

\subsection{BLAS Optimization of DNN Training Primitives}

Many Machine Learning and Deep Learning workloads can be computationally expressed in terms of Basic Linear Algebra Subroutines (BLAS), 
and Matrix Multiplication (MM) in particular \cite{8114708}.
The problem of BLAS optimization for Deep Learning applications is the target of several works concerning server-side applications.
Approximate methods can be employed to speed up the computation of MM kernels. In this context, Osawa~et~al.~\cite{8035076} proposed to apply Low-Rank Approximation to reduce the computational burden on convolutions in server applications. With their approach, they have shown up to $25 \times$ performance improvement on the wide matrices of server applications while maintaining a negligible accuracy loss.
The cost of MM kernels in Convolutional and Fully-Connected layers can be reduced by means of software approaches like the Strassen algorithm \cite{10.1145/369028.369096}. 
With this aim, Tschannen~et~al.~\cite{pmlr-v80-tschannen18a} modified the Strassen algorithm, introducing a method capable of 
learning fast approximations of MM algorithms for the end-to-end execution of DNNs.
With their approach, the authors claimed a 99.5\% reduction in the total number of multiplications in image classification models without accuracy drops.

Hardware-Software solutions are common in the acceleration of computationally-intensive tasks. In the effort to optimize General Matrix Multiplication (GEMM) with multiple data precision, Moss~et~al.~\cite{10.1145/3174243.3174258} provided support for a hardware-software GEMM framework based on an Intel HARPv2 processor, which is able to accelerate DNN models like AlexNet by up to $4 \times$ using a mixed CPU+FPGA approach.
With a similar approach, Juan~et~al.~\cite{9407138} presented a multi-threaded approach of MM for Deep Learning applications. In their approach, they exploited a 16-bit integer precision and the hardware SIMD capabilities of ARM Cortex-A processor to extend the BLIS framework. Thanks to their integer vectorized approach, they obtained a 20\% speedup with respect to FP32 models like AlexNet or VGG16 while saving 25\% energy.

The acceleration of MM workloads in MCUs is often delegated to SIMD integer computation due to the hard restrictions in terms of memory, computation, and power. In this context, ARM CMSIS-NN\cite{lai2018cmsisnn} presented a method to exploit fixed-point quantization in the form of INT16 and INT8 data to accelerate convolutions by 4.6$\times$ on ARM Cortex-M processors equipped with integer SIMD hardware. Similarly, PULP-NN \cite{8965067} provided a method to accelerate integer MM kernels on multicore RISC-V MCUs, exploiting SIMD integer computation. Thanks to their approach, they showed up to 15.5 MAC/clk in INT8 format on eight parallel RISC-V cores.
Similar to many previous works, we rely on common linear algebra kernels, i.e., MM, to solve our problem but, differently from others in this context, we focus on on-device learning based on the backpropagation algorithm with a reduced-precision capable MCU system.


\section{Background}

\begin{figure}[tb]
\centerline{\includegraphics[width=.85\linewidth]{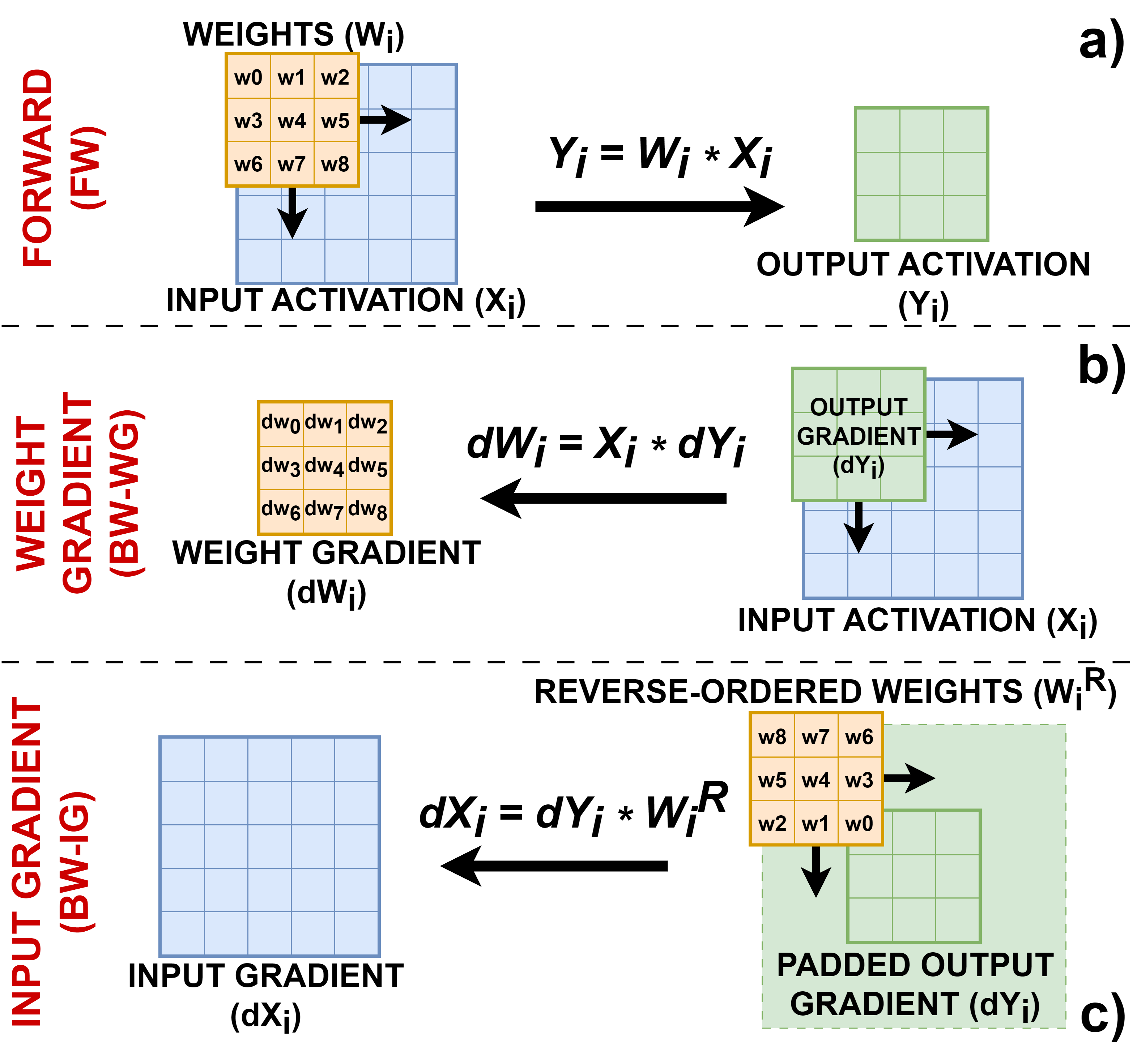}}
\caption{Training steps of a Convolution layer with a single input/output channel: a) Forward (FW) step; b) Weight Gradient (BW-WG) step, to compute the gradient of the weights; c) Input Gradient (BW-IG) step, which back-propagates the prediction error to the previous layer. We indicate as $w_i$ the filter elements. We refer to the notation introduced in \cite{9935273}.}
\label{fig.Conv2D_Theory}
\end{figure}

This section reviews the BackPropagation (BP) algorithm, a gradient-based DNN optimization technique commonly used for DNN training. 
Let us consider a DNN model composed of N layers. Every layer operates a non-linear function $ f_i(\cdot), i= 0,..,N-1 $ parameterized by the coefficient tensor $W_i$, that is learned during the training process. 

During the Forward (FW) pass, which corresponds to the DNN inference phase, the model's input data $X_0$ propagates layer-by-layer through the composite function $ \{ f_0 \circ f_1 \circ \dots \circ f_{N-1} \}$.
In the case of convolutional layers, the layer-wise operation of the FW step can be expressed as:

\begin{equation}
    \label{eq:FW}
    Y_i = W_i \ast X_i
\end{equation}
where $\ast$ denotes the cross-correlation operator (commonly denoted as \textit{convolution}), and $X_i$ and $Y_i$ are the input and output activation feature maps, respectively.
Note that $X_i \equiv Y_{i-1}$.
We omit the bias term for simplicity. 
Fig.~\ref{fig.Conv2D_Theory}--a) visually represents the FW step of a Convolution layer, operating on a single feature map. To visually describe this layer, we refer to the notation introduced in \cite{9935273}. The output of the last layer $Y_{N-1}$ represents the DNN model prediction, e.g., the class scores in case of a classification task.

To train a DNN model,  a loss function $\mathcal{L}$ is used to estimate the classification error with respect to the ground-truth labels of a set of labelled data, i.e., the train set. 
The BP algorithm has the purpose of backward propagating the prediction error to compute the gradients of the loss function with respect to the parameters of every layer's weights $W_i$, i.e., $\nabla \mathcal{L}_{W_i} = $ \textit{$dW_i$}.
Once the latter is computed, an optimization procedure, such as the Stochastic Gradient Descent (SGD), updates \textit{$W_i$} according to a certain learning rate $\eta$, e.g. \textit{$W_i$} $\leftarrow$ \textit{$W_i$} + $\eta \cdot dW_i$. 

The Backward (BW) step consists of an application of the gradient's chain rule to compute \textit{$dW_i$}. 
Starting from the network output, the gradient $\nabla \mathcal{L}_{Y_{N-1}}$ is first calculated as the derivative of the prediction error with respect to the model output. This value is backpropagated into the DNN model to compute the Intermediate Gradient (IG) tensors, denoted as \textit{$dX_i$}. As for the FW case, note that $dY_{i-1} = dX_i$.
For a convolutional layer (Fig.\ref{fig.Conv2D_Theory}--c)), the IG tensors are computed as:
\begin{equation} \label{eq:BW-IG}
    dX_i =  dY_i \ast  \delta{Y_i}/ \delta{X_i}  = dY_i \ast W_i^R
\end{equation}
This operation is referred to as BW-IG step. 
Note that, for convolutional layers,  $\delta{Y_i}/ \delta{X_i}$ corresponds to the $W_i$ tensor, opportunely transformed to $W_i^R$ by inverting the element order, as described in Sec.~\ref{sec:methods_primitives}.
The output gradient $dY_i$ may need to be padded to produce an output vector $dX_i$ with the correct size.
Following the same differentiation rule, the weight gradient $\textit{\textbf{dW}}_i $ is computed during the BW-WG step (Fig.\ref{fig.Conv2D_Theory}--b):
\begin{equation} \label{eq:BW-WG}
    dW_i =  \delta{Y_i}/ \delta{W_i} \ast dY_i  = X_i \ast dYi
\end{equation}

In the case of convolutional layers, $\delta{Y_i}/ \delta{W_i}$ is equivalent to the $X_i$ activation tensor computed during the FW pass. 

In the rest of the paper, we narrow down the scope to the workload analysis and implementation of individual layers. Hence, for simplicity of notation, we omit the index $i$ when referring to individual tensors. Tab.~\ref{tab:acronyms} summarizes the used symbols.



\begin{table}[t]
\caption{Acronyms and Symbols}
\begin{center}
\begin{tabular} { |c|c| } 

\hline
 Acronym    &    Meaning \\
\hline

 \textit{X, dX}    &  Input activation and gradient of a DNN layer \\
 \textit{W, dW}    &  Weight data and gradient of a DNN layer \\
 \textit{Y, dY}    &  Output activation and gradient of a DNN layer \\
 \textit{Im2Col}     &  Image-to-Column Operator \\ 
 \textit{Im2Row}     &  Image-to-Row Operator \\ 
 \textit{B-T}        &  Block-Transpose Operator (weights only) \\
 \textit{Tr}         &  Matrix Transposition Operator \\
 MM                  &  Matrix Multiplication \\
 MM$_T$              &  Row-Row Matrix Multiplication \\

\hline

\end{tabular}
\label{tab:acronyms}
\end{center}
\end{table}

\section{ODL kernels}
\label{sec:training_primitives}

For convolutional DNNs - i.e. DNNs whose layers mainly consist of Convolutions - the layer-wise training primitives (FW, BW-IG, BW-WG steps) reduce to convolutions as described, respectively, by Eq.~\ref{eq:FW}, \ref{eq:BW-IG}, \ref{eq:BW-WG}. Previous works~\cite{lai2018cmsisnn,8965067} showed that FW convolutions could be reshaped as Matrix Multiplications (MMs) after applying a shape transformation (e.g. \textit{Im2row} or \textit{Im2col} described below) operator to the input activation tensor. In this section, we discuss how to extend this concept to the BW steps when targeting execution on low-end MCUs. 
It is important to remark that we consider a batch size of 1 for the training task, which is equivalent to computing the weight gradients in a sample-by-sample streaming fashion.

As a template for most DNN operators, we consider a 2D Convolution (Conv2D\footnote{We refer to the Pytorch's \textit{Conv2d} notation.}) layer with weight shape $C_O \times C_I \times k_h \times k_w$, where $C_O$ is the number of output channels, $C_I$ is the number of input channels, and  $k_h \times k_w$ is the spatial filter  size.
Input and output activations (and gradients) are 3-dimensional tensors, featuring two spatial dimensions ($H$ and $W$) and a channel dimension ($C_I$ channels for $X$ and $dX$ and $C_O$ channels for $Y$ and $dY$). 
We consider the two commonly used data layouts, denoted as CHW and HWC, that differ for the ordering of the tensor dimensions in memory. The CHW convention presents the channel size as the outermost dimension, while the HWC convention stores the channel dimension as the innermost. In a matrix form, a CHW tensor is reshaped as a matrix with C rows and $H \times W$ column, where elements in a row are stored contiguously in memory. A HWC tensor is obtained by transposing the CHW matrix, i.e., a $(H \times W) \times C$ sized matrix. 
On the other hand, weight tensors (and gradients) are 4-dimensional tensors, featuring $C_O$ as the outermost dimension. The spatial filter sizes $k_h$ and $k_w$ and the input channels $C_I$ are shuffled in the inner dimensions according to the chosen layout: in case of HWC, the filter sizes are sorted as ($C_O, k_h, k_w, C_I$), while in CHW as ($C_O, C_I, k_h, k_w$).
Although input, weight, and output tensors of an individual layer could feature different memory layouts, in the following we only consider homogeneous schemes where all tensors are formatted as HWC or CHW. 
%

%

\begin{figure}[tb]
\centerline{\includegraphics[width=.5\textwidth]{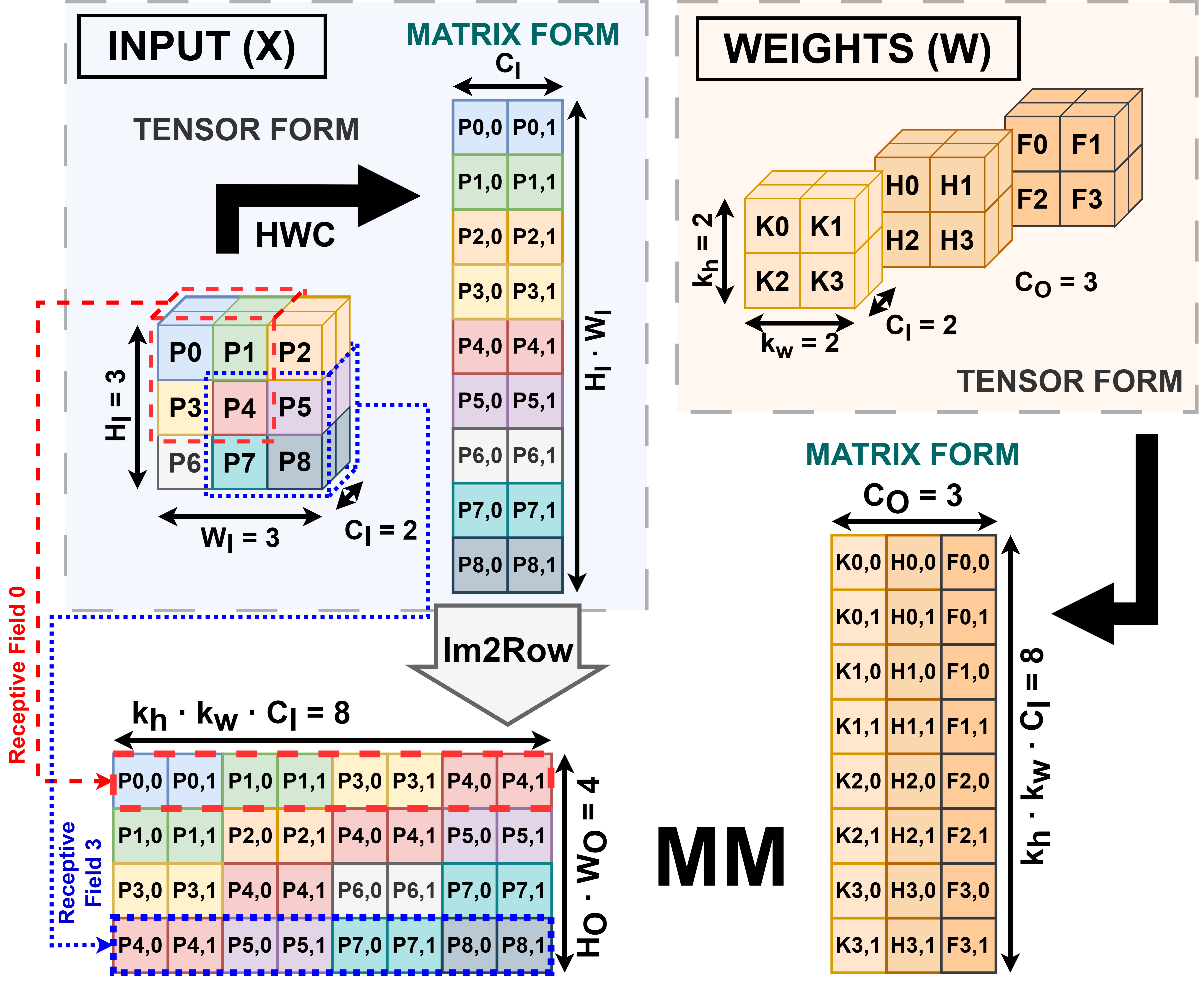}}
\caption{Matrix representation of a FW step of a Conv2D layer with with HWC data layout. The input size is $H_I=3$, $W_I=3$, $C_I=2$. The weight tensor is $k_w = k_h = 2$, $C_i=2$ and $C_o=3$.  The input tensor is transformed with \textit{Im2Row} before performing the Matrix Multiplication.}
\label{fig:Matrix_Im2Col}
\end{figure}

Fig.~\ref{fig:Matrix_Im2Col} analyzes in detail the FW step of a Conv2D with HWC layout.
The input tensor $X$ is initially stored in matrix form. Then, the \textit{Image-to-Row (Im2Row)} shape transform function copies the values under the moving window of the convolution filter (of size $k_h \times k_w \times C_I$) to a new matrix \textit{Im2Row}(X) of size $(H_O \times W_O) \times (k_h \times k_w \times C_I).$ 
The result of the convolution is then computed by means of a Matrix Multiplication between the \textit{Im2Row}(X) matrix and the weight tensor, also stored in a matrix form. 
Differently from $X$, the weight tensor $W$ is stored in matrix form by placing the elements of each filter in the columns, with adjacent $C_O$ elements.

More in detail, the \textit{Im2Row} copies data chunks of size $k_w \times C_I$ from the matrix $X$ with HWC layout to the destination matrix. On the contrary, with an input featuring a CHW layout, the chunk size of the \textit{Im2Row} data transfer is reduced to the $k_w$ elements that are stored contiguously in memory. A strided access is performed to load the next elements that fall under the weight filter. The different memory layouts are key for impacting the efficiency of the \textit{Im2Row} transform function, as discussed in the experimental section and, in particular, when the datatype is set to FP16.  

\begin{figure}[t]
\centerline{\includegraphics[width=.5\textwidth]{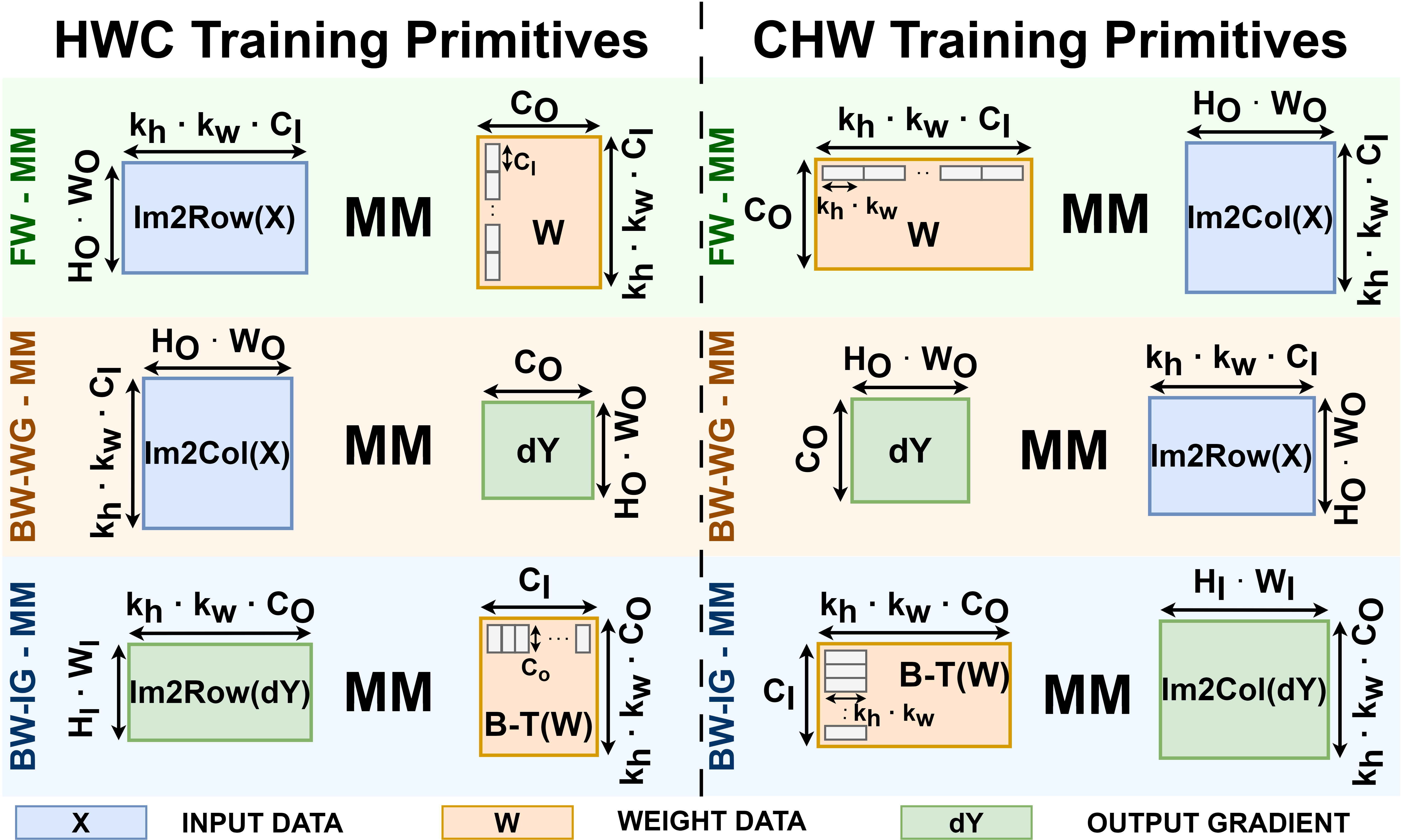}}
\caption{ODL training primitives of a Conv2D Layer. On the left, the non-optimized HWC expressions of the Conv2D training primitives for ODL; on the right, the same expressions in CHW format.}
\label{fig.Conv2d_HWC_CHW}
\end{figure}

\begin{figure}[t]
\centerline{\includegraphics[width=.45\textwidth]{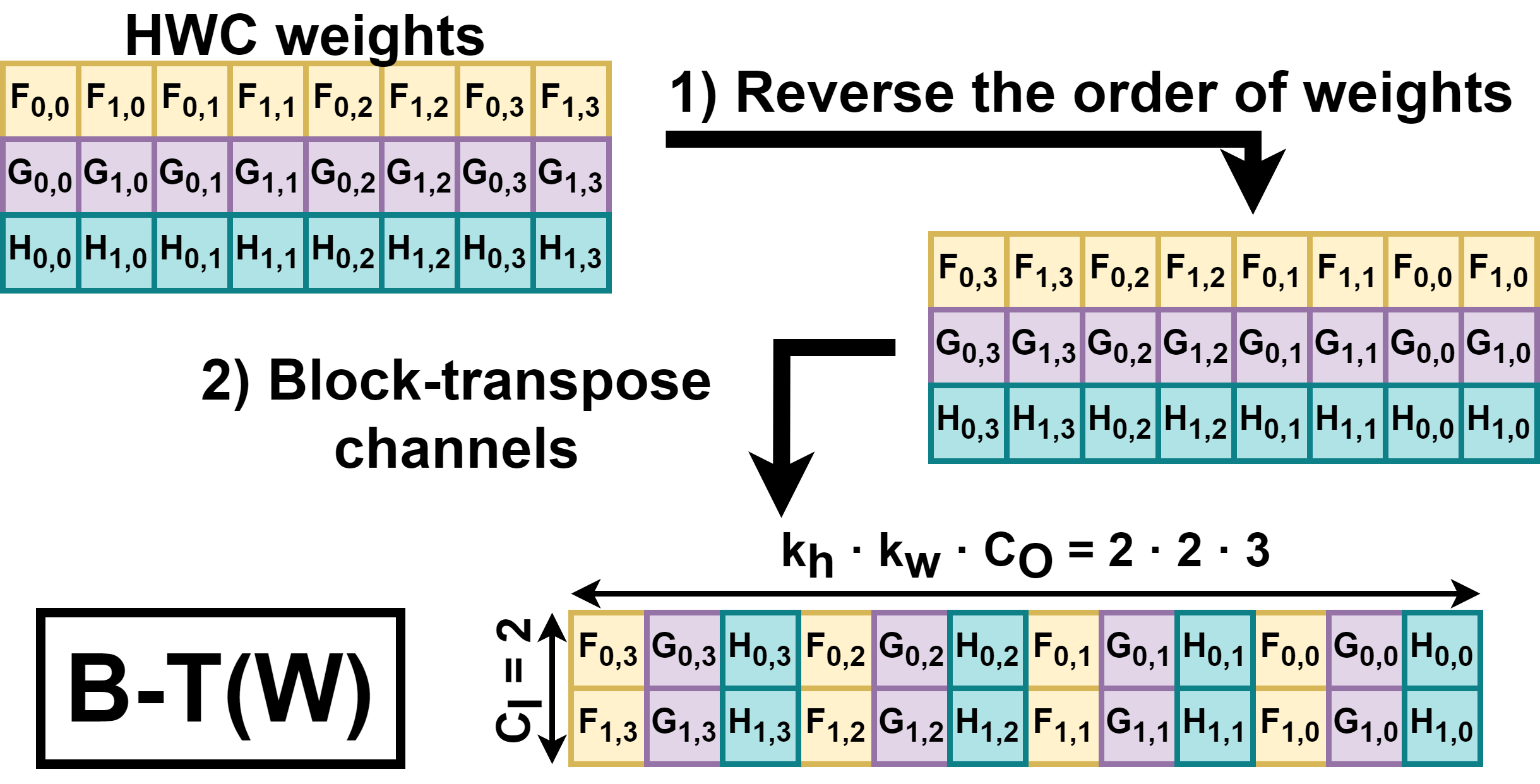}}
\caption{Workflow of the \textit{Block-Transpose} (\textit{B-T}) operator applied to a HWC weight matrix of size $3 \times 2 \times 2 \times 2$.}
\label{fig.Tr_B-T}
\end{figure}

In case of CHW, the expression of the FW step is adapted, as shown in the top of Fig.~\ref{fig.Conv2d_HWC_CHW}, to produce a transposed output matrix with respect to the HWC case. Differently from the HWC expression: (i) operands are transposed and (ii) the MM switches the operand order. 
To handle the transposition of the input activation $X$, the \textit{Im2Row} operator is replaced with the \textit{Image-to-Column (Im2Col)} operator, where elements under the filter are copied on a column of the destination matrix. In terms of performance, similar considerations to the ones drawn for \textit{Im2Row} also hold for \textit{Im2Col}. In general, the resultant \textit{Im2Row}($X$) (or \textit{Im2Col}($X$)) matrix features a memory footprint larger than $X$ because every element of the input tensor contributes to the computation of multiple output values. 
The memory requirements of an \textit{Im2Row/Im2Col} exceeding the available memory can be reduced using tensor tiling - i.e., instead of processing the full tensor, multiple partial sub-tensors can be copied in sequence in a temporary buffer, using the transform operators, and processed at minimal computation overhead with respect to processing the full tensor \cite{9381618}.

In addition to the FW step of the Conv2D layer, Fig.~\ref{fig.Conv2d_HWC_CHW} visually shows the core operations of BW training primitives operating on tensors with HWC e CHW layouts.
In the plot, we denote the Matrix Multiplications, which implement each training step as FW-MM for the forward step and BW-WG-MM and BW-IG-MM for the backward steps. 
Similarly to the FW, the BW-WG convolution is turned into a Matrix Multiplication to compute the weight gradient $dW$. 
This step takes as inputs the activation input $X$, stored after the FW pass, and the gradient vector $dY$. 
Note that the $dY$ tensor has the same size of $Y$. 
Differently from the FW-MM, the \textit{Im2Col} transform is applied over the $X$ tensor with an HWC layout. \textit{Im2Row}($X$) is instead used for CHW. 
Lastly, the BW-IG step is reshaped into a Matrix Multiplication  (BW-IG-MM) between the output gradient $dY$ and the weight tensor $W$. 
Differently from the other steps, the BW-IG step requires the weight tensor to be transformed using a \textit{Block-Transpose} (\textit{B-T}) operator before feeding the BW-IG-MM.
Fig.~\ref{fig.Tr_B-T} illustrates the workflow of the \textit{B-T} operator applied to a weight matrix $W$. First, the $k_h \times k_w$ elements of the filters are placed in reverse order, i.e., the reversed-order matrix $W_i^R$ of Eq.~\ref{eq:BW-WG}; 
second, the weight input and output channels are \textit{block-transposed}: elements belonging to the same input channel are transposed into rows.

\begin{figure}[t]
\centerline{\includegraphics[width=.5\textwidth]{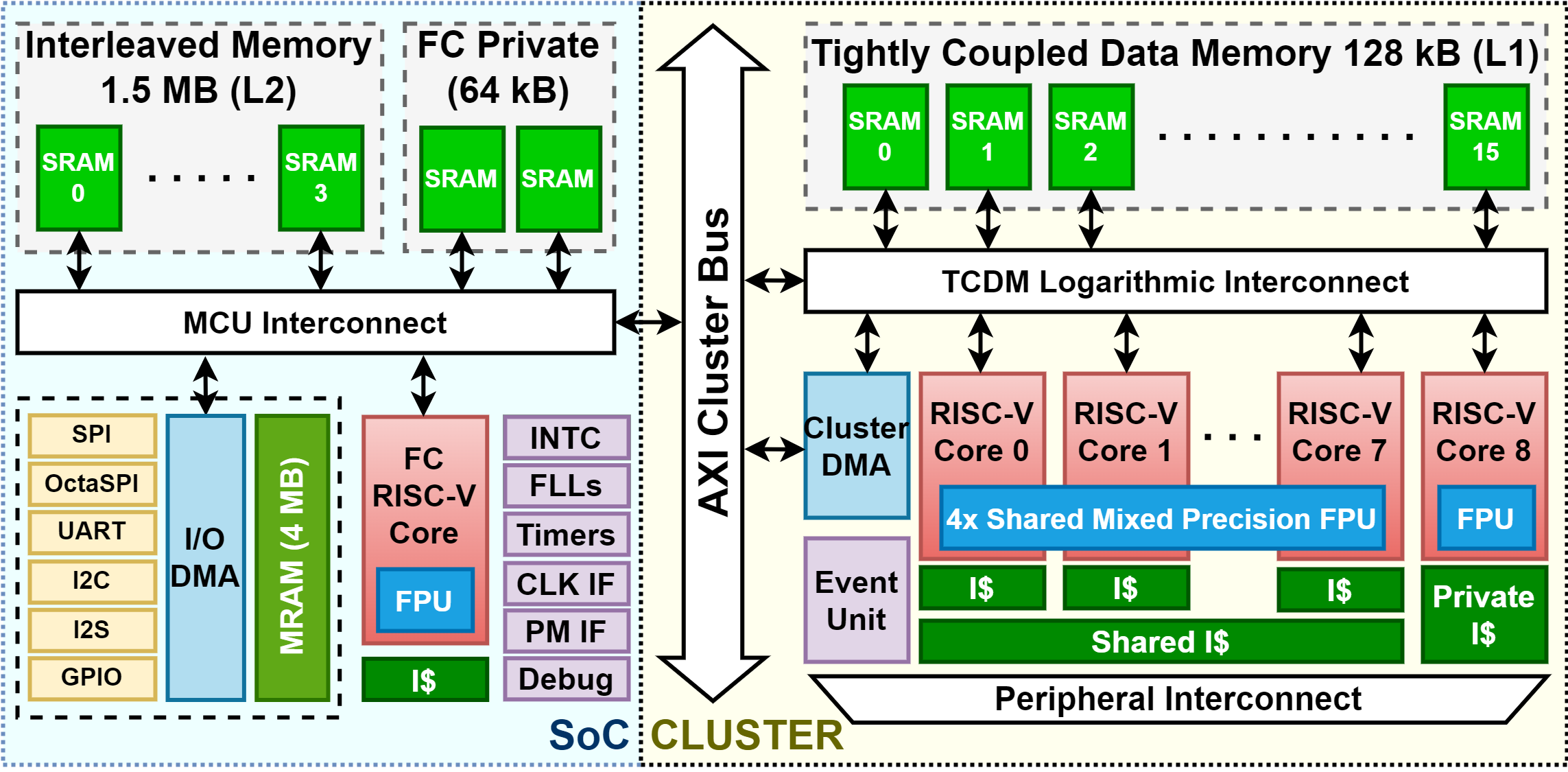}}
\caption{PULP SoC Architecture with 8 RISC-V Cores. The PULP Cluster is equipped with 4 shared Mixed Precision Floating Point Units (FPUs) to compute FP32 and FP16 operations.}
\label{fig:PULP-SoC}
\end{figure}

\section{ODL on a MultiCore MCU with FP16 support}
\label{sec.OptTheory}

This section describes our design methodology for latency-optimized  ODL software kernels  targeting a multi-core platform with HW support for reduced-precision FP16 SIMD instructions.

\subsection{The PULP Platform} 
Fig.~\ref{fig:PULP-SoC} shows the RISC-V-based Parallel Ultra-Low Power (PULP) platform targeted by our approach~\cite{rossi2021vega}, as embodied in Greenwaves GAP9 SoC. 
The system features an MCU domain, namely the PULP SoC region (depicted in light blue in the figure), which includes a single RISC-V core for control-related tasks, and a Cluster domain (in yellow) with 8+1 RISC-V cores to accelerate compute-intensive tasks. 
All the cores support the RV32IMFC ISA, extended with DSP-oriented instructions, like post-increment load/store instructions and 2-level hardware loops. 
Every CPU is also granted access to a mixed-precision Floating Point Unit (FPU), operating full-precision (FP32) and half-precision (FP16) floating-point instructions.
More in detail, every FPU can process  1$\times$ FP32 MAC in a single clock cycle or 2  MAC/clk if using FP16 SIMD instructions.

From a system-level viewpoint, the PULP SoC features a multi-level memory hierarchy with up to 2 MB of L2 SRAM, directly accessible by the MCU core in a single clock cycle, and an on-chip non-volatile MRAM memory of up to 4MB. 
On the Cluster side, an L1 data-scratchpad memory with a size of up to 256 kB is shared among the multiple cores. 
The Cluster DMA can be used to efficiently copy data between the L1 and the L2 memories in the background of the CPUs operation. 
Data in the L1 memory can be accessed in a single clock cycle by the cluster cores.

In our setup, we consider the 8+1-core PULP Cluster for the acceleration of the DNN training primitives.  
Out of the total 9 cores, the first 8 are devoted to parallel computation and can access 4 shared mixed-precision FPUs. 
The 9-th core, instead, acts as a Cluster Controller: this core is in charge of programming the Cluster DMA and dispatching parallel tasks to the other 8 compute cores.

\begin{figure}[tb]
\centerline{\includegraphics[width=.45\textwidth]{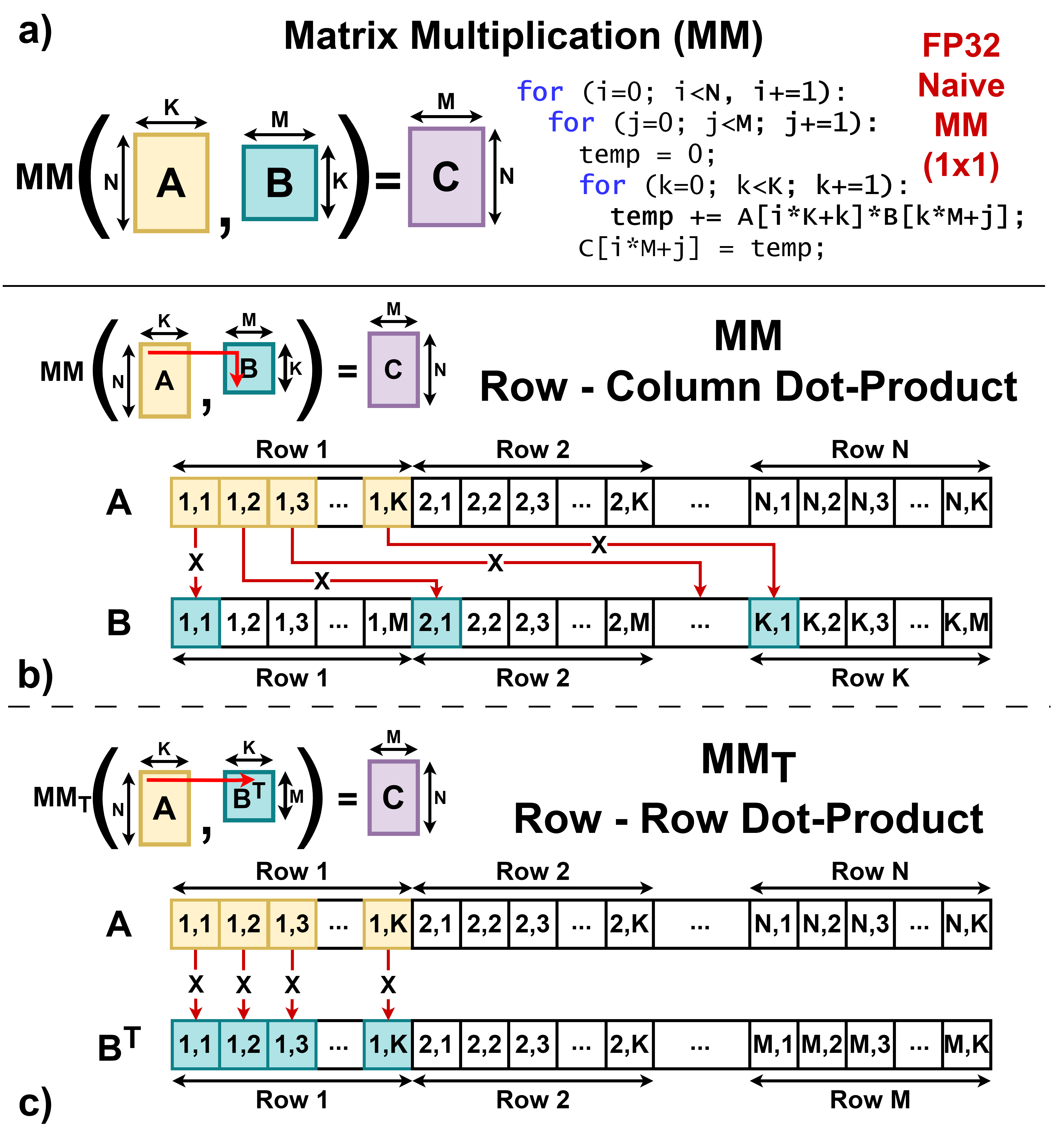}}
\caption{a) Pseudo-code of the FP32 non-unrolled ($1 \times 1$) na\"{i}ve MM algorithms that we refer to as baseline; b) and c) show the difference between the memory access patterns of a MM (b) and MM$_T$ (c) kernels. MM$_T$ favours SIMD loads and MAC instructions, as $B$ matrix elements are adjacent in memory. Row elements of the matrices are stored adjacent, while column elements have a stride equal to the row length one each other.
}
\label{fig.MMOptFP16}
\end{figure}

\subsection{Matrix Multiplication Optimization}

As highlighted in Section~\ref{sec:training_primitives}, 
the MM algorithm is the computation core of the ODL primitives. 
Therefore, we first study the acceleration of the MM on the targeted platform using either FP32 or FP16 datatypes. 

Let us consider a generic matrix multiplication with $\textit{A} \in\mathbb{R}^{N\times K}$ and $\textit{B} \in\mathbb{R}^{K\times M}$ inputs and  $\textit{C} \in\mathbb{R}^{N\times M}$ as output.
\textit{A} and \textit{B} are stored in memory as arrays. The elements of a row (K in case of \textit{A}) are adjacent in memory. Conversely, successive column elements (N in case of \textit{A}) are stored with a stride equal to the row length.
Fig.~\ref{fig.MMOptFP16}--a) shows the pseudo-code of a na\"{i}ve MM implementation that uses 3 nested \texttt{for} loops and a time complexity of $\mathcal{O}(N \times K \times M)$. 
For every iteration of the inner loop, the CPU operates a MAC between elements loaded from the \textit{A} and \textit{B} arrays. 
Our baseline implementation assumes data stored in low-level memory, i.e., the L1 memory of the PULP cluster.
Hence, elements from $A$ and $B$ arrays are loaded in a single clock cycle by the Cluster cores.

Fig.~\ref{fig.MMOptFP16}--b) shows the memory access pattern to compute the dot product between a row vector of matrix $A$ and a column vector of matrix $B$. 
While the elements from a row of the matrix $A$ are accessed from a contiguous memory area, strided accesses are required to load the elements belonging to a column of the matrix $B$.
If we consider that every element is a FP16 number, the access pattern of this \textit{Row-Column Dot-Product} is inefficient in loading columns, as it cannot use 32-bit load/store instructions to load two FP16 elements in one single-clock instruction.
This motivates us to consider an MM$_T$ operator that expects the second operand $B$ in a transposed form according to:

\begin{equation}
MM(A, B) = MM_T(A, Tr(B))
\label{eq.SIMDOpt}
\end{equation}
where $Tr$() is the transpose operator. 
Differently from the MM baseline, the MM$_T$ performs a series of dot-products between the row vectors of \textit{A} and $Tr$(B). 
%
%
As a major benefit, this Row-Row Dot Product scheme, which is depicted in \ref{fig.MMOptFP16}--b), gains a sequential memory access pattern by design both for the matrix $A$ and matrix $B$, favoring the usage of SIMD FP16 load/store instructions.
%
On the other side, the transposition of the $B$ matrix represents a potential computation overhead. 
However, this extra cost can be cancelled by transposing  the $B$ matrix before the deployment on the target platform, when possible, e.g., transposing the matrix of weight values. 
This cost can also be absorbed by the shape transform operator of the ODL training primitives, i.e., by replacing an \textit{Im2Col} function with an \textit{Im2Row} or vice versa. This strategy will be discussed in further detail in Section~\ref{sec:methods_primitives}.

\begin{figure}[tb]
\centerline{\includegraphics[width=.5\textwidth]{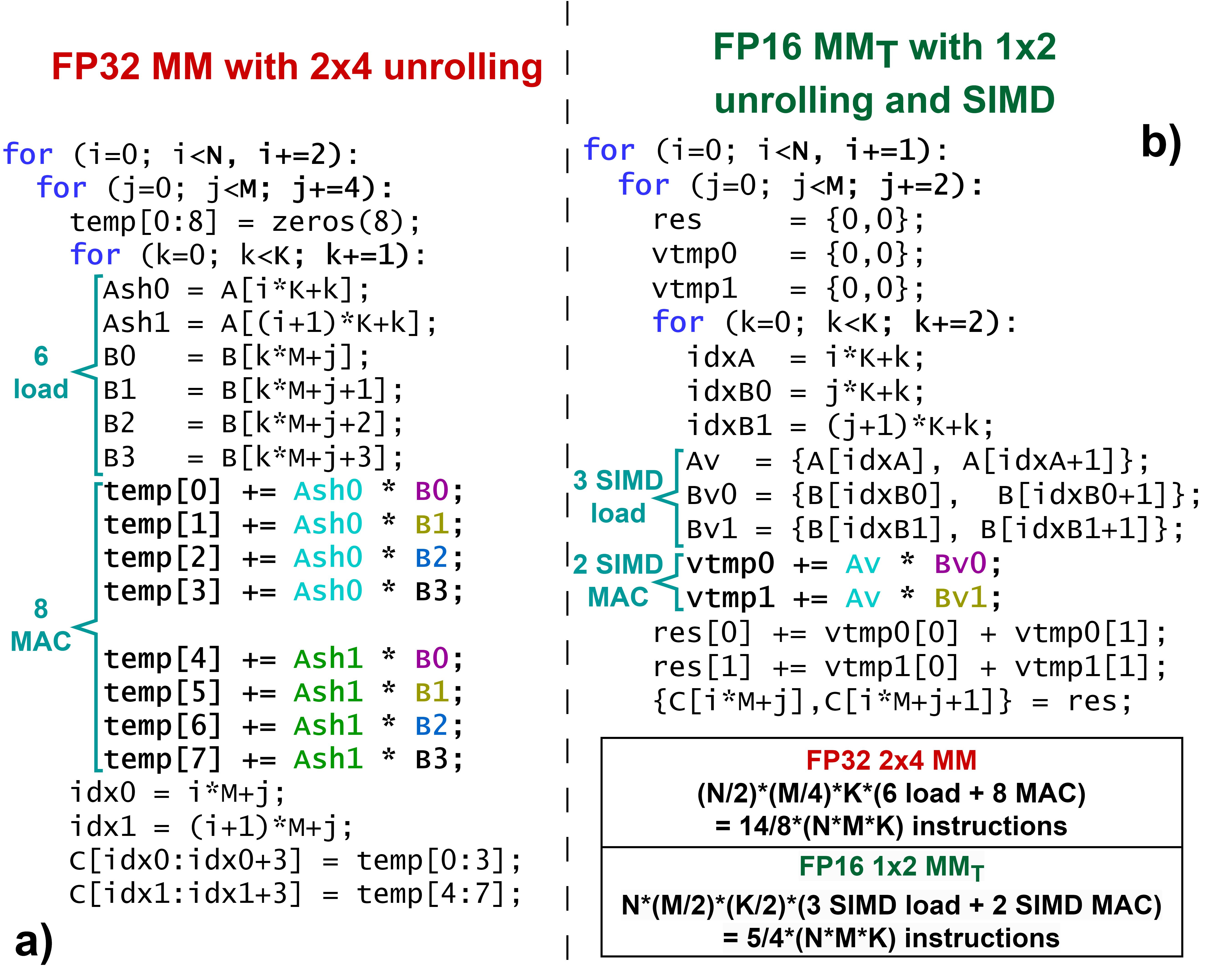}}
\caption{Pseudo-code of optimized MM algorithms: a) shows an FP32 MM with $2 \times 4$ unrolling; b) presents an FP16 $1 \times 2$ MM$_T$ which makes use of SIMD to load adjacent $A$ and $B$ row elements. While performing the same amount of MAC instructions per iteration, SIMD MM$_T$ allows to reduce by 43\% the inner loop instructions.
}
\label{fig.MMOptFP16_2}
\end{figure}

\subsection{Loop Unrolling and Parallelization}
As proposed in~\cite{10.1007/978-3-031-15074-6_13}, we exploit loop unrolling and parallelization to speed up the MM and MM$_T$ kernels. 
Loop unrolling maximizes the data reuse of loaded elements.  
We refer to an unrolling factor of \textit{U} $\times$ \textit{V} to indicate a MM kernel that computes $U\times V$ elements of the output matrix $C$ within the inner loop.
A MM with a higher unrolling factor presents a lower number of instructions by using fewer load operations.
If the MM dimensions are not divisible by the unrolling factors, ancillary \textit{leftover} loops take care of the remainders using a na\"ive non-unrolled strategy.

Fig.~\ref{fig.MMOptFP16_2}--a) shows the pseudo-code of an FP32 MM with a $2 \times 4$ unrolling. 
This kernel computes 8 partial results of the output matrix $C$ within the innermost loop, obtained from the $2 \times 4$ MAC operations.
In this case, the CPU loads only 6 values instead of 16 because every element from the $A$ and $B$ arrays is reused 4 and 2 times, respectively.
Hence, the utilization of the MAC units in the innermost loop increases from 33\% to 57\% if compared to a na\"{i}ve implementation (Fig.~\ref{fig.MMOptFP16}--a)
- i.e. the number of MAC instructions in the inner loop is increased in exchange for less load instructions.
Fig.~\ref{fig.MMOptFP16_2}--b) shows the pseudo-code of an FP16 MM$_T$ exploiting 1$\times$2 loop unrolling. 
Thanks to the SIMD instructions, the inner loop computes 4 MAC at the cost of 3 load operations, reaching the same MAC utilization of the previous FP32 kernel but with a lower unrolling factor.

Lastly, we exploit the multi-core architecture of the PULP Cluster and the native parallelism of the MM computation. 
The parallelization strategy  that we adopt splits the iterations  of the outermost loop dimensions with respect to the available cores (8 in our case). 
Thanks to this, the execution throughput gains a parallel speedup that increases almost linearly with the number of parallel cores.
For instance,  a na\"{i}ve FP32 MM with $32 \times 32$-shaped matrices parallelized on 8 RISC-V cores shows a parallel speed-up of up to 7.47 vs. a theoretical limit of 8.

\begin{figure}[tb]
\centerline{\includegraphics[width=.45\textwidth]{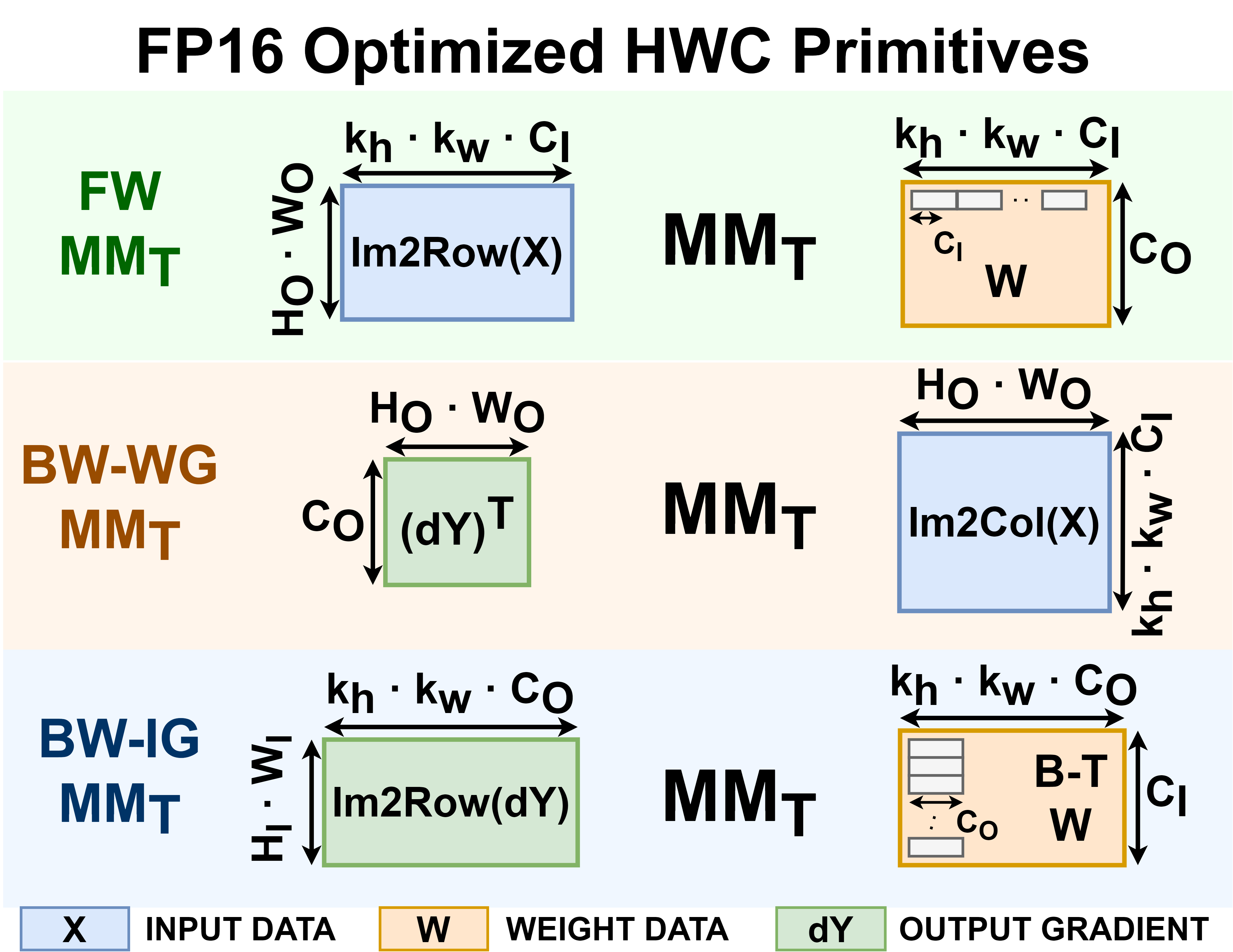}}
\caption{ODL matrix expressions of the FP16 training primitives of a Conv2D, which make use of the MM$_T$ kernels.
}
\label{fig.Conv2d}
\end{figure}

\begin{table*}[t]
\centering
\footnotesize
\caption{
On-Device Learning operations for FP32 and FP16 Conv2D and PW Conv layers. $A$ and $B$ denote the first and second operand of the used MM kernel. $W^T$ indicates a tensor already transposed in memory. \textit{Tr(W)} transpose instead the operand at runtime.
}
\resizebox{\textwidth}{!}{%
\begin{tabular}{|l||c|c|c||l||c|c|c|}
\hline

\textbf{Conv2D - HWC}  &  MM Kernel & $A$ ($1^{st}$ Operand) & $B$ ($2^{nd}$ Operand) & \textbf{PW Conv - HWC}    &  MM Kernel & $A$ ($1^{st}$ Operand) & $B$ ($2^{nd}$ Operand) \\
\hline
\textit{FP32 FW}       &            &  \textit{Im2Row}(X)  &       W            &   \textit{FP32 FW}          &            &      X       &      W          \\
\textit{FP32 BW-WG}    &      MM    &  \textit{Im2Col}(X)  &       dY           &   \textit{FP32 BW-WG}       &     MM     &      Tr(X)   &      dY          \\
\textit{FP32 BW-IG}    &            &  \textit{Im2Row}(dY) &     B-T(W)         &   \textit{FP32 BW-IG}       &            &      dY      &      Tr(W)        \\
\hline
\textit{FP16 FW}       &            &  \textit{Im2Row}(X)  &       W$^T$        &   \textit{FP16 FW}          &              &    X       &      W$^T$        \\
\textit{FP16 BW-WG}    &   MM$_T$   &  Tr(dY)              & \textit{Im2Col}(X) &   \textit{FP16 BW-WG}       &    MM$_T$    &    Tr(dY)  &      Tr(X)        \\
\textit{FP16 BW-IG}    &            &  \textit{Im2Row}(dY) &    B-T(W$^T$)      &   \textit{FP16 BW-IG}       &              &    dY      &      Tr(W$^T$)     \\
\hline
\hline

\textbf{Conv2D - CHW}  &  MM Kernel & $A$ ($1^{st}$ Operand) & $B$ ($2^{nd}$ Operand) & \textbf{PW Conv - CHW} &   MM Kernel & $A$ ($1^{st}$ Operand) & $B$ ($2^{nd}$ Operand) \\
\hline
\textit{FP32 FW}       &            &   W                  & \textit{Im2Col}(X)   & \textit{FP32 FW}       &            &    W     &   X     \\
\textit{FP32 BW-WG}    &     MM     &    dY                & \textit{Im2Row}(X)   & \textit{FP32 BW-WG}    &   MM       &    dY    &   Tr(X)   \\
\textit{FP32 BW-IG}    &            &    B-T(W)            & \textit{Im2Col}(dY)  & \textit{FP32 BW-IG}    &            &  Tr(W)   &   dY    \\
\hline
\textit{FP16 FW}       &            &      W               & \textit{Im2Row}(X)   &   \textit{FP16 FW}     &            &     W    &   Tr(X)  \\
\textit{FP16 BW-WG}    &    MM$_T$  &      dY              & \textit{Im2Col}(X)   &   \textit{FP16 BW-WG}  &   MM$_T$   &     dY   &   X     \\
\textit{FP16 BW-IG}    &            &   B-T(W)             & \textit{Im2Row}(dY)  &   \textit{FP16 BW-IG}  &            &   Tr(W)  &  Tr(dY)  \\
\hline

\end{tabular}
}
\label{tab:SIMDDataMarsh}
\end{table*}

\subsection{FP16 ODL Primitives}
\label{sec:methods_primitives}
The design of the FP16 ODL primitives is based on the software templates of the PULP-TrainLib \cite{10.1007/978-3-031-15074-6_13}.
As this library only included CHW FP32 training kernels, we \textit{i)} extended it with the shape transform operators for both HWC and CHW layouts and \textit{ii)} introduced FP16 primitives. 
In the remainder of the discussion, we focus mainly on the Conv2D case, but similar considerations and design strategies have been applied to the main other layers composing typical DNNs.

Fig.~\ref{fig.Conv2d} graphically depicts our  FP16 ODL training primitives for a Conv2D layer with an HWC data layout that exploits the  MM$_T$ kernels. 
Differently from FP32, we replace the MM with MM$_T$ to fully benefit from row-by-row SIMD Dot Products.
To amortize the computational cost of the transpose operator required by Eq.~\ref{eq.SIMDOpt}, we store the weight parameters of our HWC primitives in transposed form to comply with the layout required by MM$_T$ kernels. 
This choice impacts the BW-WG step: the produced weight gradient $dW$ must also be transposed for a convenient update of the weight tensor. 
For this reason, we further transpose the BW-WG-MM expression and feed \textit{Tr(\textit{dY})}. 
This additional transform has a negligible impact on the execution costs since it brings a latency overhead lower than 5\%.

Tab.~\ref{tab:SIMDDataMarsh} provides a summary of the operations for the FP32 and FP16 ODL training primitives. We consider  Conv2D and PointWise Layers.
More in detail, the table highlights the transforms and MM kernels for the FW, BG-WG, and BW-IG steps when an HWC or a CHW layout is used. Differently from the FP32 Conv2D operations that use MM kernels (reported in Fig.~\ref{fig.Conv2d_HWC_CHW}), the FP16 Conv2D implementations make use of MM$_T$ and shape transform functions to transpose the $B$ operand. 

Unlike other cases, the primitives for FP16 HWC require the weights to be stored in memory in a transposed form (see Eq. \ref{eq.SIMDOpt}).
In the table, we denote this weight tensor as $W^T$. 
Because of this transformation, the HWC FW and BW-IG steps feature the same order of operands of the FP32 counterpart in the convolution expression. 
Conversely, the operands of the FP16 BW-WG step are switched to produce a transposed weight gradient, which can be directly summed to $W^T$ during the weight update phase. 
On the contrary, FP16 Conv2D CHW primitives do not require to store the weights in transposed form. In this case, the transposition of $B$ can be obtained, instead, by replacing each \textit{Im2Row} operator with \textit{Im2Col} and vice versa.

Despite the fact that we only discussed the Conv2D case, our ODL kernel design methodology can be used for every convolutional DNN layer. 
As a notable example, Tab.~\ref{tab:SIMDDataMarsh} also lists the internal operations for the training steps of a Pointwise (PW) Convolution, which is frequently used for DNN models, e.g., DepthWise Separable layers~\cite{https://doi.org/10.48550/arxiv.1704.04861}.
Given a filter size of $ k_w = k_h = 1$, the ODL steps do not include any \textit{Im2Col} or \textit{Im2Row} transforms. 
On the contrary, the weights are transposed during the BW-IG step. Similarly to Conv2D primitives, the FP16 HWC primitives require to store transposed weights and to swap the operands in the BW-WG step. CHW primitives, instead, only transpose the B operand. Furthermore, an additional transposition is required in the FP16 CHW FW step, unlike HWC.


\section{Experimental Results}
\label{sec.results}

\begin{figure*}[tb]
\centerline{\includegraphics[width=\textwidth]{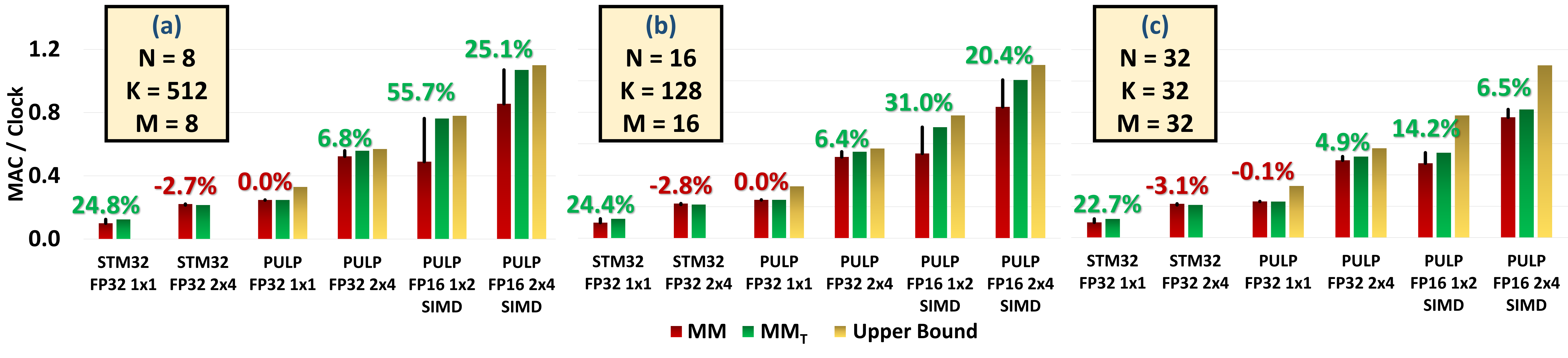}}
\caption{MAC/clk of different MM kernels on both an STM32L476RG and 1 RISC-V core. All cases perform 32768 MAC. Thanks to vectorization, we achieve a top 25.17\% performance gain. However, the structure of the FP16 MM kernels limits the performances if K is small due to the accumulation of the inner product introduced by the intermediate loop.}
\label{fig.MMSU}
\end{figure*}

\subsection{Implementation Details}

We evaluate our software design  on a RISC-V-based Multi-Core MCU, Greenwaves Technologies' GAP9. This platform embodies an instance of the PULP Platform with a 9-core Cluster equipped with 4 shared FPUs and HW support for SIMD FP16 instructions. 
%
In our implementation, input (output) data and gradients are stored in the large off-cluster L2 memory and are copied to (from) L1 using the Cluster DMA before (after) the computation of each training step.
The \textit{Im2Col} and \textit{Im2Row} transform functions are operated by the cluster cores to gain data load/store parallelism; the destination matrix, also placed in the L1 memory, feeds the MM kernel.  
When the FP16 datatype is used, the amount of data to be copied is reduced by 2$\times$ with respect to FP32, leading to faster shape transform operators. 
Padding can introduce a large overhead because of the extra additional control instructions located in the inner loop of the copy to check for for zero-insertion. 
For example, an \textit{Im2Col} operator applied to an input activation of size $8 \times 8 \times 1$ may suffer up to a $60\%$ cycle increase in case of a $3 \times 3$ filter. Larger input channel sizes help reducing this overhead.


\subsection{FP16 and FP32 Optimized Matrix Multiplications}

First, we study the performance of the optimized MM kernels on the targeted platform. 
Fig.~\ref{fig.MMSU} shows the throughput expressed as a ratio between the amount of MAC and the measured clock cycles, i.e., MAC/clk. An increase of MAC/clk score corresponds to a faster execution.
We analyze single-core runs of multiple-sized MM and MM$_T$ functions featuring different unrolling factors for both FP16 and FP32 datatypes.
For every setting, we report the upper bound limit (yellow bar), accounted by excluding from the cycle count any instructions but MAC, loads, and stores. 
For comparison purposes, we also benchmark our FP32 MM kernels on an STM32L4 MCU.

The FP32 MM baseline without loop unrolling (marked as $1\times 1$  in the plot) presents an average throughput of 0.24 MAC/clk for all three considered cases.  
The same performance is measured for the transposed form MM$_T$ because the latency spent to access and process 32-bit data is the same. 
This is up to $2.4 \times$ faster than the same kernel running on the STM32L4 device, 
%
thanks to the build tools, which fully exploit the underlying hardware by leveraging post-increment load/store instructions and hardware loops to reduce the iteration overheads.
Using $2 \times 4$ loop unrolling, the throughput of MM kernels further increases by $2.11 \times$ with respect to the baseline.
Compared to an equivalently unrolled STM32L4 porting, this is also $2.36 \times$.
Further increasing the unrolling factor is detrimental: the register file pressure of unrolling requires frequent register spilling in the stack, leading to severe slow-down effects.

Introducing FP16 SIMD vectorization and MM$_T$ kernels, a maximum speed-up of  1.91$\times$ is measured vs. the fully unrolled FP32 case, reaching a top performance of 1.07 MAC/clk.
For both the analysed unrolling factors, the performance varies across the matrix sizes but is generally superior to FP16 MM kernels.
When the innermost K dimension is large (e.g., Fig.~\ref{fig.MMSU}-a), the MM$_T$ with $2\times 4$ loop unrolling shows a $\sim$25\% performance gain, which leads to performance near the theoretical upper bound.
This condition is common in the deep layers of convolutional models, featuring a relatively large number of channels (e.g., more than 8 input and output channels and a spatial size of 8).
When K is small, the gain is less substantial but still present ($\sim$6\% in Fig.~\ref{fig.MMSU}-c).
The reason for this effect is that innermost loops with more iterations positively impact the execution of unrolled MM and MM$_T$ kernels, amortizing the overhead introduced by the result accumulation and store instruction of the outer loops.

\subsection{Conv2D ODL Primitives}

In this section, we evaluate the primitive-level optimizations introduced in Sec.~\ref{sec.OptTheory}. 
Tab.~\ref{tab:conv} reports the shapes of the four Conv2D layers and a PointWise layer under analysis. The sizes are chosen as portions - or tiles - of several input and weight tensors that fit layers from the ResNet8 model in Fig.~\ref{fig.ResNet8}. 
In particular, CONV1, CONV2 and CONV3 are possible tiles of Layer 2 and 3, featuring large $H \times W$ size and $C$ size. Conversely, CONV4 is a possible tile of the input layer, while PW CONV represents Layer 9 of the ResNet8 model. In these experiments, FP32 layers use the MM kernels, while FP16 ones employ MM$_T$. A HWC data layout is adopted.

\begin{figure*}[t]
\centerline{\includegraphics[width=\textwidth]{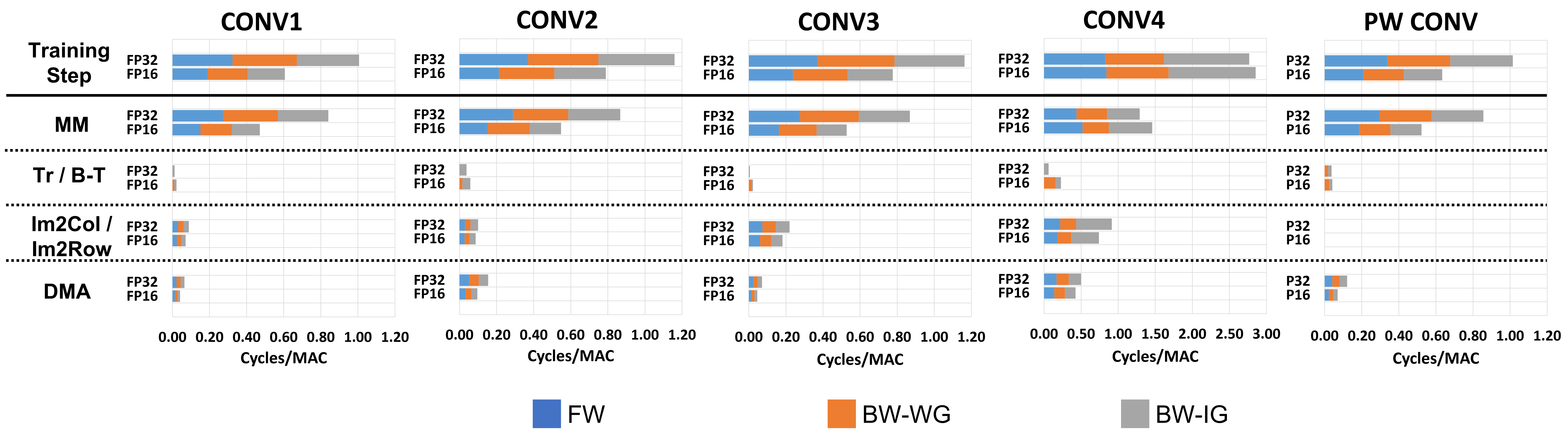}}
\caption{ Training Latency (top row) of four 2D Convolution layers, namely CONV1--4 and a PointWise Convolution layer, namely PW CONV, whose shapes are reported in Tab.~\ref{tab:conv}. Measurements are expressed in terms of cycles/MAC. A latency breakdown of the total performances is provided in the bottom rows (lower results are faster). }
\label{fig.Conv2DOpt}
\end{figure*}

Fig.~\ref{fig.Conv2DOpt} shows the latency for a complete training, which consists of the FW and BW steps with respect to a single data point, of the considered layers when leveraging 8-core processing. 
Measurements are shown in terms of normalized latency, measured in clock cycles, with respect to the total number of MAC (cycles/MAC). A lower score indicates a smaller latency.
On the top row, we highlight the latency breakdown among the FW, BW-IG, and BW-WG phases, while on the bottom rows, we provide a detailed report of the internal operations: MM/MM$_T$ kernels, \textit{Tr/B-T} transpose operators, \textit{Im2Col/Im2Row}, and the DMA transfers between the L1 and L2 memories. 

For every layer shape, the training step is dominated by the MM kernels in a fully-unrolled version. When applied to Conv2D training, FP32 $2 \times 4$ MM kernels achieve up to 3.66 MAC/clk when executing both FW and BW-IG steps of CONV1. 
In the same case, FP16 $2 \times 4$ SIMD MM$_T$ kernels achieve 6.63 MAC/clk, outperforming by $1.81 \times$ the FP32 kernels on 8 parallel cores.
A similar result has already been observed for the single-core case (Fig.~\ref{fig.MMSU}). 

In case of a single FW or BW step, FP16 SIMD optimizations of the $MM_T$ kernel achieve up to $1.72 \times$ performance increase, as observed for the CONV1, CONV2, CONV3 and PW CONV scenarios.
This speedup is uniform across all the training steps. 
Only for CONV4, the usage of FP16 SIMD brings a slower execution than the FP32 computation. 
In this corner case, both the innermost loop and the external loops feature a reduced size because of the single channel of the input image, preventing the kernel from exploiting acceleration opportunities given by the loop unrolling. 
Instead, a leftover subroutine is invoked to handle the operation, slowing down the process. 
This infrequent case may appear in the first layer of a model; a kernel without loop unrolling or using FP32 should be preferred in this case. 
However, when considering a full model design, e.g., ResNet8, this type of layer has a very limited impact on the on the total computation time, i.e., less than 3\% on a ResNet8.

The percentage of latency related to MM and MM$_T$ kernels with respect to the total latency depends on the layer's size.   
In particular, a larger channel size increases the weight tensor size and, therefore, the impact of DMA transfers. 
On the contrary, large H and W sizes increase the computation intensity. 
These effects result in a total execution latency that is dominated by more than 76\% on average by MM/MM$_T$ kernels with shapes like CONV1. 
In other cases like CONV2 and CONV3, instead, MM/MM$_T$ kernels impact the 68\% of the latency due to increased DMA activity (CONV2, due to larger weight tensor size compared to CONV1) and shape transformation (CONV3, due to larger activation tensor size) overheads. These extra costs determine a slight decrease in the performance of the ODL primitives, whose latency is increased by up to 11\% even with large channel and spatial sizes. 
The throughput of the MM kernels is substantially reduced, as already observed, in the case of small channel sizes, like for CONV4. In the case of a PW CONV, the compute efficiency reaches 4.76 MAC/clk in FP16 on average, 4.5\% less than CONV1; the MM/MM$_T$ represents 84\% of the total latency.

The execution of the Conv2D ODL primitives is also highly influenced by the \textit{Im2Col} and \textit{Im2Row} operators, which can represent a large overhead during each training step. 
The impact of these shape transform operators is particularly relevant when the sizes of the input and output tensors widely exceed the size of the weight tensor. In the case of CONV3 and CONV4 with FP32 data type, these overheads represent on average the $19\%$ and $33\%$ of the total latency cost, respectively. On the contrary, layers with smaller input and output tensor sizes are less impacted by the compute cost of these shape transforms. 
The latency due to \textit{Im2Col/Im2Row} operators is reduced by $1.2 \times$ thanks to the FP16 SIMD. 
This depends on the capability to move two contiguous data elements with a single 32-bit load/store. 
This speedup is almost constant for each layer shape, including the corner cases as CONV4.
On the other side, \textit{Im2Col/Im2Row} operators are not used by the PointWise layer PW CONV because of the $1 \times 1$ filter size.

\begin{table}[t]
\centering
\footnotesize
\caption{Tile Shapes of ResNet8 Conv2D and PointWise Layers }
\begin{tabular}{c||c|c|c|c|c|c|c|c}
\hline
\textbf{Layer} & \textbf{$C_I$} &  \textbf{$H_I$} &  \textbf{$W_I$} & \textbf{$k_h$} & \textbf{$k_w$}  &  \textbf{$C_O$} &  \textbf{$H_O$} &  \textbf{$W_O$} \\ 
\hline \hline

CONV1        & 16 & 8  & 8  & 3  & 3  & 16 & 8  & 8  \\ \hline
CONV2        & 16 & 4  & 4  & 3  & 3  & 32 & 4  & 4  \\ \hline
CONV3        & 8  & 16 & 16 & 3  & 3  & 8  & 16 & 16 \\ \hline
CONV4        & 1  & 8  & 8  & 3  & 3  & 16 & 8  & 8  \\ \hline 
PW CONV      & 32 & 8  & 8  & 1  & 1  & 64 & 8  & 8  \\ \hline
\hline
\end{tabular}
\label{tab:conv}
\end{table}

In the baseline FP32 implementation of Conv2D primitives, the \textit{B-T} operator typically represents less than 3\% of the total latency of a training step, depending on the amount of input and output channels. FP16 optimizations introduce additional transposition operators in the ODL primitives, to fully exploit MM$_T$ kernels (Tab.~\ref{tab:SIMDDataMarsh}). This extra overhead is however limited to 4\% at most in typical layer and tile sizes, reaching $8\%$ only for CONV4 where the impact of the MM/MM$_T$ reduces.
In case of PW CONV, no \textit{B-T} operator is required (Tab.~\ref{tab:SIMDDataMarsh}). Extra latency costs are however accounted for the transpose operators, reaching up to $4\%$ and $6\%$ of the total workload for, respectively, the FP32 and FP16 format.

Lastly, DMA data transfers between L2 and L1 memory represent 7\% of the execution time in typical tile shapes, like CONV1 and CONV3. 
Given the large channel size, CONV2 features an increased weight tensor size with respect to CONV1 and CONV3. This increases the DMA latency, which reaches up to 14\% of the total time. 
DMA transfers may occupy up to 20\% in corner cases like CONV4. 
When FP16 is used, the time to transfer data is reduced by $1.6 \times$ with respect to FP32 on average, thanks to the halved memory footprint. In case of PW CONV, both FP32 and FP16 DMA transfers are responsible for at most the 12\% of the total latency, representing the prime overhead.


\subsection{Energy Evaluation}

\begin{figure}[t]
\centerline{\includegraphics[width=.45\textwidth]{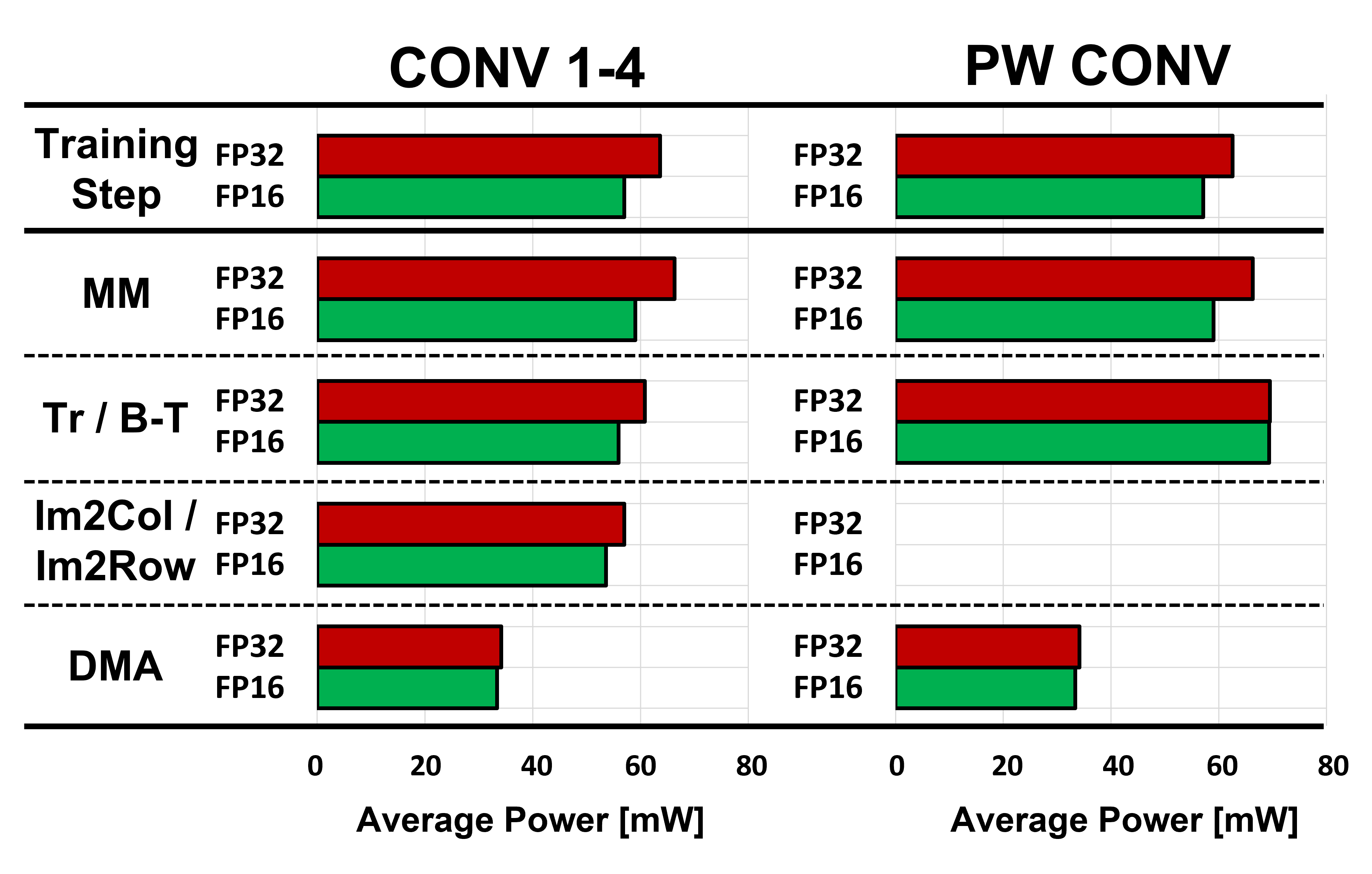}}
\caption{ Average Power analysis of CONV 1-4 and PW CONV, whose latency analysis was presented in Fig.~\ref{fig.Conv2DOpt}. Results are presented in mW. } 
\label{fig.PowerBreakdown}
\end{figure}

In this section, we evaluate the energy consumption of the layer primitives of Tab~\ref{tab:conv}. To this aim, we measure the power consumption of the building components of the training steps on a GAP9 SoC featuring a supply voltage of 0.8 V and a running clock frequency of 370 MHz. 
The energy profile is then calculated by taking into account the latency of the primitives shown in Figure~\ref{fig.Conv2DOpt}.
Fig.~\ref{fig.PowerBreakdown} shows the power costs (in $mW$) of CONV 1-4 and PW CONV. We break down the contributions from the different components (MM, DMA and the transform operators) either for the FP32 and FP16 kernels. The power consumption is plotted after averaging the measurements across the training steps; a low variance was observed because of the similar workload composition. 

The average power consumption of CONV 1-4 reaches up to 63.6 mW for both FP32 and FP16. 
This is a result of the prominence of MM/MM$_T$ operators (70\% of the total latency), whose consumption surpasses 66 mW in FP32 and 59 mW in FP16.
When it comes to FP16, the MM$_T$ kernels have a lower power consumption, suggesting that the hardware is not being fully utilized due to a reduction of the parallel efficiency of FP16 MM$_T$ kernels. 
The maximum speedup is limited to 6.23 on 8 cores. 
This is due to the matrix shapes of each training step in FP16, which does not offer favorable parallelization schemes.
%
%
This effect can also be observed in terms of MAC/clk: if a theoretical MAC/clk of 7.13 is expected ($2 \times$ the FP32 performance on the same layer), only a 6.37 MAC/clk is measured on the FP16 primitives. 
This $12\%$ difference is  reflected in the decrease of power consumption.
%
%

The FP32 and FP16 \textit{Im2Row/Im2Col} operators feature an average power consumption of 55.3 mW. 
The similar cost of these operators indicates a similar activity of the hardware units. 
%
In the case of PW CONV, the average power consumption approximates 60 mW, in line with CONV 1-4. 
In this case, the power consumption due to \textit{Tr/B-T} entirely depends on transposition operators, which consume 69.4 mW on average independently of the datatype. 
%
Both CONV 1-4 and PW CONV feature similar power consumption for both MM kernels and DMA transfers. These latter represent the smallest power overhead, as their consumption is as little as 33.5 mW on average. 

When analyzing the energy consumption, we account for a minimum of 4 $\mu J$ in the case of a full FP32 training step of CONV4, and a maximum of 25.4 $\mu J$ for CONV1. 
In the case of FP16, the range goes from 3.8 $\mu J$ for CONV4 (corner case with a latency similar to FP32) and 17.4 $\mu J$ for CONV3.
Reflecting the latency breakdown observed in Fig.~\ref{fig.Conv2DOpt}, the transform operators (i.e., \textit{Im2Col/Im2Row} and \textit{Tr/B-T}) of CONV 1-3 reach up to the 18.6\% of the single training steps in FP32 and 27.3\% in FP16. 
The same operators in the CONV4 consume up to the 47\% of the total, because of the reduced channel size of the input activation, which makes the BW-IG \textit{Im2Row/Im2Col} more impacting and limit the maximum performance of MM/MM$_T$ kernels.

\begin{figure}[tb]
\centerline{\includegraphics[width=0.45\textwidth]{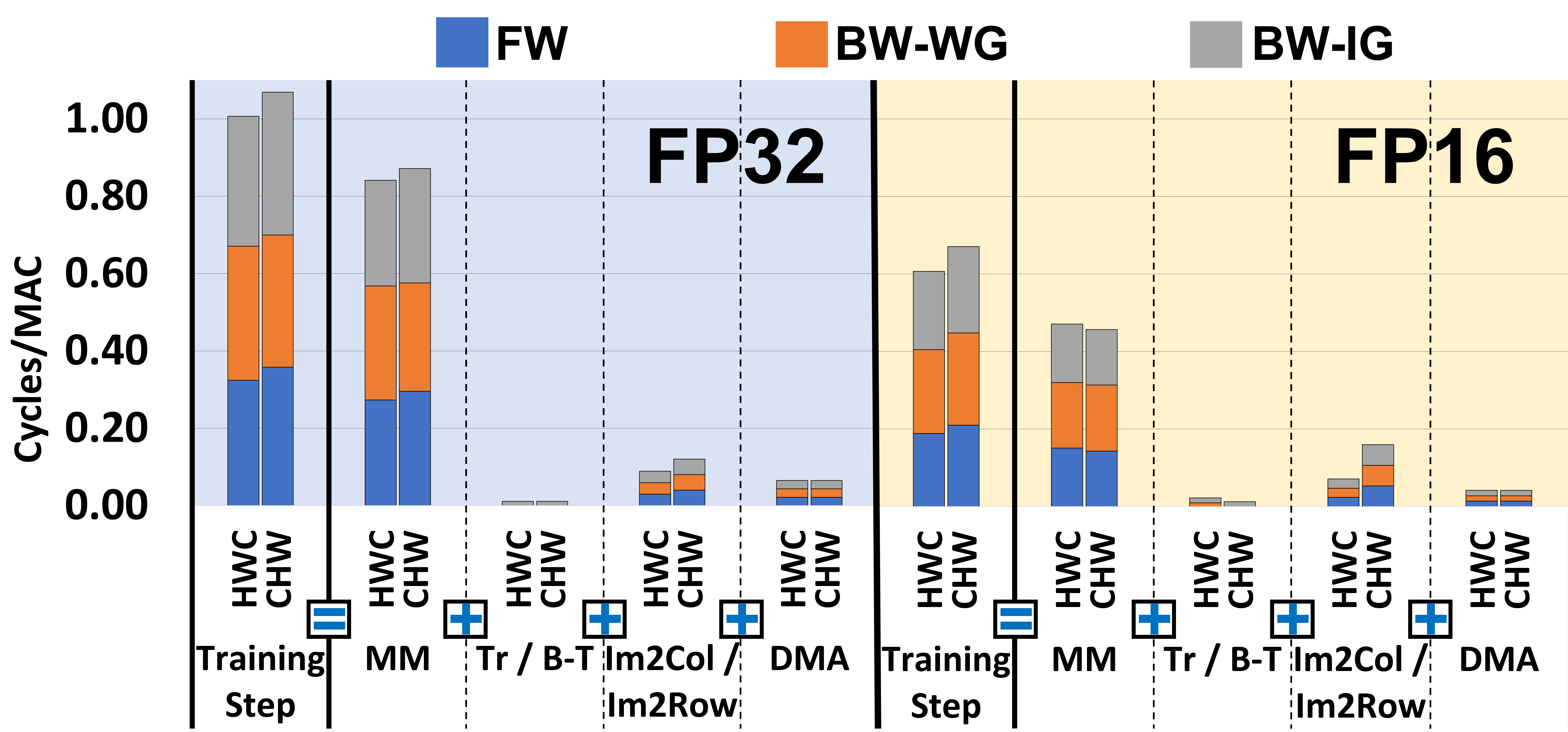}}
\caption{ Comparison of CHW and HWC formats for a Conv2D layer of the same size with both FP32 and FP16. The considered Conv2D has the same shape of Fig.~\ref{fig.Conv2DOpt}'s CONV1. }
\label{fig.HWCvsCHW}
\end{figure}

\subsection{Conv2D HWC/CHW Layout Comparison}
Fig.~\ref{fig.HWCvsCHW} shows a comparison of the performances achieved with both HWC and CHW formats on the same Conv2D shape (CONV1).
In the case of FP32 training primitives, MM kernels are on average responsible of the largest share of the execution time for each training step, both in the case of CHW primitives (81\%) and of HWC primitives (83\%).
FP32 DMA transfers and transpositions present the same latency since data can be accessed and manipulated element-by-element, disregarding vectorization. 
However, the contribution coming from the \textit{Im2Col/Im2Row} operators is reduced by 36\% using the HWC format. 
This depends on the structure of HWC \textit{Im2Col} and \textit{Im2Row} algorithms, which load and reshape $k_h \times k_w \times C_i$ tensor elements in each iteration of the inner loop. 
On the contrary, CHW operators manipulate only $k_h \times k_w$ elements for every iteration, leading to larger reshaping overheads due to additional control instructions. 
Furthermore, the HWC format allows slightly faster execution of MM kernels in some cases. 
This depends on the shape of the involved matrices, which can allow larger sizes on the N dimension. In turn, this may enable larger chunks when parallelizing outer loops on multiple cores. 
Overall, FP32 HWC-shaped Conv2D kernels show up to 6\% faster latency with respect to CHW.

In the case of FP16 primitives, MM$_T$ kernels occupy 68\% of the latency of CHW primitives. 
The dominance of MM$_T$ kernels is largely increased in the case of HWC kernels since MM$_T$ occupies 78\% of the latency.
The impact of DMA transfers is the same in both FP32 and FP16. 
In terms of $MM_T$ kernels, both CHW and HWC formats have similar performance  since both formats provide similar matrix shapes in each training step. 
In the HWC case, FP16 \textit{Im2Col} and \textit{Im2Row} kernels achieve 2$\times$ faster performance than in the CHW case, thanks to fully vectorizable data accesses (see Fig.~\ref{fig:Matrix_Im2Col}), impacting the overall training step latency only by 12\%. 
In the CHW case, transfer chunks are smaller than in HWC, yielding larger overheads impacting as much as 24\% the overall latency.
In absolute terms, CHW FP16 \textit{Im2Col/Im2Row} algorithms suffer a 32\% slowdown with respect to FP32 due to poorer vectorization. Instead, the same FP16 HWC operators prove to be $1.25 \times$ faster than FP32.
Therefore, FP16 SIMD execution in HWC data format proves to be 11\% faster than CHW in the same data precision.

\begin{figure}[tb]
\centerline{\includegraphics[width=0.5\textwidth]{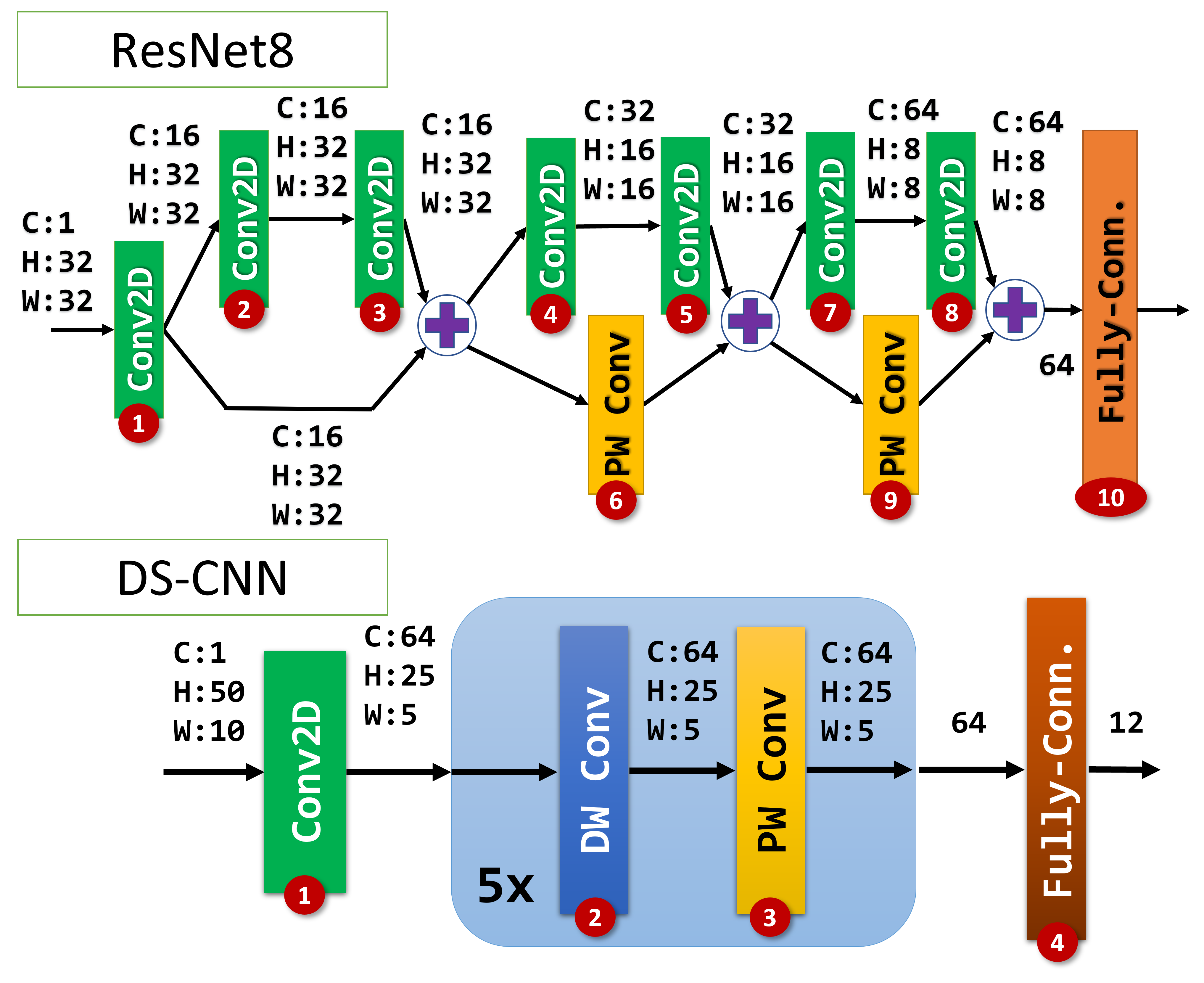}}
\caption{ResNet8 (top) and DS-CNN (bottom) model architectures including the shapes of the activation tensors and layer types.}
\label{fig.ResNet8}
\end{figure}

\begin{table}[t]
\centering
\caption{Latency on GAP9 for a complete training step of ResNet8 and DS-CNN models and comparison with top inference-only scores on MCUs.  }
\resizebox{\linewidth}{!}{%
\begin{tabular}{l|cccc}
\hline
                                       & \multicolumn{2}{c|}{ResNet 8}    & \multicolumn{2}{c}{DS-CNN} \\ \hline
\textbf{GAP9 Training {[}This work{]}} & FP32 & \multicolumn{1}{c|}{FP16} & FP32         & FP16        \\ \hline
Total Clock Cycles {[}Millions{]}      & 11.9 & \multicolumn{1}{c|}{6.3}  & 4.3          & 2.4         \\
\% MM                                  & 0.73 & \multicolumn{1}{c|}{0.74} & 0.71         & 0.75        \\
latency GAP9@240MHz {[}ms{]}           & 49.5 & \multicolumn{1}{c|}{26.3} & 17.7         & 9.9         \\
latency GAP9@370MHz {[}ms{]}           & 32.1 & \multicolumn{1}{c|}{17.1} & 11.5         & 6.4         \\ \hline
\end{tabular}
}
\label{tab:end-models}
\end{table}

\subsection{ End-To-End TinyML Model Training}

In this section, we apply the proposed methodology to the layers of complete TinyML models - a ResNet8 for Image Classification and a DS-CNN for Audio Keyword Spotting, whose architectures are represented in Fig.~\ref{fig.ResNet8}.
For both models, we estimate the latency for running a training step on a single input sample (i.e., batch size of 1) using our FP16 or FP32 primitives. 
We use an HWC format to implement the model layers; only DepthWise Convolutions are represented in memory using a CHW layout. 

From a memory viewpoint, we consider storing the activation tensors, the weights and their gradients in the on-chip L2 RAM memory. 
The total memory footprint of the FP32 ResNet8 amounts to 893 kB and reaches 772 kB in the case of the DS-CNN model. 
These costs are reduced by $2 \times$ if the FP16 datatype is used, dropping to 443 kB and 386 kB, respectively.
On the other side, the FP32 and FP16 versions maximize the utilization of the L1 memory buffer, which we limit to 64 kB for storing the tensor tiles. 
E.g., In the case of ResNet8, the computation of the FP32 FW step of the Layer 2 can be split into 32 tiles, each one featuring a memory cost of 47 kB. 
Thanks to the 2x compression of FP16, the number of channels per tile is doubled: the workload splits into 16 tiles with a requirement of 53 kB per tile. 
A similar tiling logic is proposed for the other layers and the DS-CNN layers.

Tab.~\ref{tab:end-models} shows the latencies measured on GAP9 in terms of clock cycles. 
In the case of the ResNet8 model, the overall training time is dominated by Conv2D layers (up to 98\%) for both FP32 and FP16 formats. 
The Pointwise layers, instead, represent up to 2.3\% of the workload.
Using the FP16 primitives accelerates the training time by 1.88$\times$ vs. the FP32 implementation. 
Yet, the MM operator presents the largest execution time, up to 74\% of the total training time. 
Moreover, these kernels show high efficiency (4.2 MAC/clk on average for MM in Resnet8) because large tile size can be considered, similarly to CONV1 and CONV2 of Fig.~\ref{fig.Conv2DOpt}.
The \textit{Im2Row/Im2Col} shape transforms (up to 14\% for FP16), and the DMA transfers (7\%) constitute the other most expensive tasks.
On the other side, the workload of the DS-CNN is dominated by the 5$\times$  Depthwise Separable Convolutions, and in particular, the PointWise layers, which take more than 54\% of the computation in both FP32 and FP16.

Running on GAP9 clocked at 370MHz, a complete training step of ResNet8 and DS-CNN takes, respectively, 17.1 msec and 6.4 msec.  
This result paves the way for real-time ODL on low-end MCUs using the canonical backpropagation algorithm. 
Considering a data streaming scenario, the ResNet8 and the DS-CNN models can sustain the throughput of, respectively, image sensor acquisition (typically 10-30 fps for embedded applications) and audio frame processing every 0.5s used for Keyword Spotting. 
This also holds when GAP9 works in low-power mode, 240MHz@0.65V, but the average power consumption of the training task reduces to 27.18 mW, $2.23\times$ lower than working at full speed. 
Given the existing slack between the training latency and typical sensor data rate, we speculate on the feasibility of a real-time mini-batch-based ODL framework that we will investigate in future work.

\subsection{Continual Learning on MCU case-study}

Lastly, we evaluate our design in a class-incremental Continual Learning approach proposed by Pellegrini et. al.~\cite{pellegrini2020latent}. 
A MobileNet128 trained on a 10-class image classification task learns to recognize a new class after acquiring 100 image samples of the new objects, without forgetting previously learned classes. The new data are labelled and mixed with 500 pre-stored embeddings of the other classes. 
This data set is used to  fine-tune the last layers of the MobileNet128 for learning the 11-th class; an SGD optimization is applied over 8 epochs.

Table~\ref{latency-table} compares our solution with other State-of-the-Art ODL frameworks to solve the task. 
More in detail, we estimate the latency and the energy consumption to train the weight parameters of the last 7 layers (up to layer \textbf{DW21}), the last 5 layers (\textbf{DW23}), or only the last layer (\textbf{LIN27}). 
For comparison purposes, we consider the most mature ODL software libraries for MCU:  AIfES~\cite{AIfES}, targeting an STM32L476RG, and PULP-TrainLib~\cite{10.1007/978-3-031-15074-6_13}, which targets multi-core RISC-V MCUs but only supports FP32 ODL kernels.

When fine-tuning multiple layers, our solution demonstrates to be up to 1.6$\times$ and 767$\times$ faster than PULP-TrainLib on GAP9 and AIfES on an STM32L4 MCU, respectively. 
While the first is motivated by the shift from FP32 to FP16, the second large gap depends on multiple factors: the different ISA and CPU micro-architecture ($\sim$2.4$\times$), the reduced precision ($\sim$1.6$\times$), the 8-core parallelization ($\sim$7.5$\times$), the GAP9 clock frequency higher than STM32 ($\sim$4.6$\times$) and our HW/SW optimization that we estimate to contribute  $\sim$6$\times$.
Conversely, the gain is reduced when retraining only the last layer. 
FP16 fully-connected training primitives are not yet fully optimized with SIMD vectorization; in this case, our solution achieves the same performance level as the FP32 implementation of PULP-TrainLib. 
Moreover, our design is 292.02 $\times$ more energy efficient compared to AIfES; the lower gain compared with the performance speedup is due to GAP9's higher power consumption than an STM32L4 at maximum speed.
Thanks to the proposed technique, the adaptation time for highly accurate Continual Learning solutions~\cite{pellegrini2020latent} is reduced from whole days (entirely infeasible!) to 3-5 minutes, depending on the embedding layer depth (\textbf{DW21} or \textbf{DW23}).

\begin{table}[t]
\centering
\caption{Latency and Energy Evaluation for Continual Learning on MCUs}
\resizebox{\columnwidth}{!}{%
\begin{tabular}{c|c|cc|cc}
\hline
\multicolumn{1}{l|}{\textbf{}}           & \multicolumn{1}{c|}{\textbf{Platform}} & \multicolumn{2}{c|}{\textbf{Latency (s)}} & \multicolumn{2}{c}{\textbf{Energy (J)}} \\ \hline
\multirow{3}{*}{\textit{Aifes}\cite{AIfES}}          & \multirow{3}{*}{STM32L4 @ 80 MHz}      & \textbf{DW21}           & 236655          & \textbf{DW21}           & 7739         \\
                                         &                                        & \textbf{DW23}           & 142157          & \textbf{DW23}           & 4649          \\
                                         &                                        & \textbf{LIN27}          & 102.4           & \textbf{LIN27}          & 3.35          \\ \hline

\multirow{3}{*}{PULP-TrainLib\cite{10.1007/978-3-031-15074-6_13}} & \multirow{3}{*}{Greenwaves GAP9 @ 370 MHz}            & \textbf{DW21}           & 504.1           & \textbf{DW21}           & 32.24         \\
                                         &                                        & \textbf{DW23}           & 303.8           & \textbf{DW23}           & 19.43         \\
                                         &                                        & \textbf{LIN27}          & 0.89            & \textbf{LIN27}          & 0.06          \\ \hline

\multirow{3}{*}{This Work}     & \multirow{3}{*}{Greenwaves GAP9 @ 370 MHz}       & \textbf{DW21}           & 308.5           & \textbf{DW21}           & 18.68          \\
                                         &                                        & \textbf{DW23}           & 185.9           & \textbf{DW23}           & 11.26          \\
                                         &                                        & \textbf{LIN27}          & 0.89            & \textbf{LIN27}          & 0.06          \\ \hline
\end{tabular}
}
\label{latency-table}
\end{table}


\section{Conclusion}

In this paper, we introduced a novel methodology to optimize the execution of DNN primitives for On-Device Learning on multi-core MCU powered by FP16 SIMD FPUs. 
We proposed a strategy to reshape the training kernels into Matrix Multiplication operators. 
Furthermore, we provided an efficient implementation of the MM kernel exploiting FP16 SIMD instructions and we defined the needed transform functions for the different steps of the training algorithm. 
When benchmarked on a Continual Learning scenario, our solution deployed on a multi-core RISC-V MCU was demonstrated to be more than two orders of magnitude faster than the other proposed ODL library for single-core MCUs.
To foster future research on MCU-based On-Device Learning, we release the code of our library as open-source software at: \url{https://github.com/pulp-platform/pulp-trainlib}.



\begin{thebibliography}{10}
\expandafter\ifx\csname url\endcsname\relax
  \def\url#1{\texttt{#1}}\fi
\expandafter\ifx\csname urlprefix\endcsname\relax\def\urlprefix{URL }\fi
\expandafter\ifx\csname href\endcsname\relax
  \def\href#1#2{#2} \def\path#1{#1}\fi

\bibitem{10.1145/3469029}
M.~G.~S. Murshed, C.~Murphy, D.~Hou, N.~Khan, G.~Ananthanarayanan, F.~Hussain,
  Machine learning at the network edge: A survey, ACM Comput. Surv. 54~(8) (oct
  2021).
\newblock \href {https://doi.org/10.1145/3469029} {\path{doi:10.1145/3469029}}.

\bibitem{wang2020deep}
F.~Wang, M.~Zhang, X.~Wang, X.~Ma, J.~Liu, Deep learning for edge computing
  applications: A state-of-the-art survey, IEEE Access 8 (2020) 58322--58336.

\bibitem{Zemlyanikin_2019_ICCV}
M.~Zemlyanikin, A.~Smorkalov, T.~Khanova, A.~Petrovicheva, G.~Serebryakov,
  512kib ram is enough! live camera face recognition dnn on mcu, in:
  Proceedings of the IEEE/CVF International Conference on Computer Vision
  (ICCV) Workshops, 2019.

\bibitem{lamberti_tinydronet}
L.~Lamberti, V.~Niculescu, M.~Barciś, L.~Bellone, E.~Natalizio, L.~Benini,
  D.~Palossi, {{Tiny-PULP-Dronets: Squeezing Neural Networks for Faster and
  Lighter Inference on Multi-Tasking Autonomous Nano-Drones}}, in: 2022 IEEE
  4th International Conference on Artificial Intelligence Circuits and Systems
  (AICAS), 2022, pp. 287--290.
\newblock \href {https://doi.org/10.1109/AICAS54282.2022.9869931}
  {\path{doi:10.1109/AICAS54282.2022.9869931}}.

\bibitem{rusci2022accelerating}
M.~Rusci, M.~Fariselli, M.~Croome, F.~Paci, E.~Flamand, Accelerating rnn-based
  speech enhancement on a multi-core mcu with mixed fp16-int8 post-training
  quantization, arXiv preprint arXiv:2210.07692 (2022).

\bibitem{8930945}
M.~Zanghieri, S.~Benatti, A.~Burrello, V.~Kartsch, F.~Conti, L.~Benini, Robust
  real-time embedded emg recognition framework using temporal convolutional
  networks on a multicore iot processor, IEEE Transactions on Biomedical
  Circuits and Systems 14~(2) (2020) 244--256.
\newblock \href {https://doi.org/10.1109/TBCAS.2019.2959160}
  {\path{doi:10.1109/TBCAS.2019.2959160}}.

\bibitem{9278587}
D.~Brunelli, T.~Polonelli, L.~Benini, Ultra-low energy pest detection for smart
  agriculture, in: 2020 IEEE SENSORS, 2020, pp. 1--4.
\newblock \href {https://doi.org/10.1109/SENSORS47125.2020.9278587}
  {\path{doi:10.1109/SENSORS47125.2020.9278587}}.

\bibitem{banbury2020benchmarking}
C.~R. Banbury, V.~J. Reddi, M.~Lam, W.~Fu, A.~Fazel, J.~Holleman, X.~Huang,
  R.~Hurtado, D.~Kanter, A.~Lokhmotov, et~al., Benchmarking tinyml systems:
  Challenges and direction, arXiv preprint arXiv:2003.04821 (2020).

\bibitem{https://doi.org/10.48550/arxiv.1606.06565}
D.~Amodei, C.~Olah, J.~Steinhardt, P.~Christiano, J.~Schulman, D.~Mané,
  Concrete problems in ai safety (2016).
\newblock \href {https://doi.org/10.48550/ARXIV.1606.06565}
  {\path{doi:10.48550/ARXIV.1606.06565}}.

\bibitem{weiss2016survey}
K.~Weiss, T.~M. Khoshgoftaar, D.~Wang, A survey of transfer learning, Journal
  of Big data 3~(1) (2016) 1--40.

\bibitem{de2021continual}
M.~De~Lange, R.~Aljundi, M.~Masana, S.~Parisot, X.~Jia, A.~Leonardis,
  G.~Slabaugh, T.~Tuytelaars, A continual learning survey: Defying forgetting
  in classification tasks, IEEE transactions on pattern analysis and machine
  intelligence 44~(7) (2021) 3366--3385.

\bibitem{konevcny2016federated}
J.~Kone{\v{c}}n{\`y}, H.~B. McMahan, D.~Ramage, P.~Richt{\'a}rik, Federated
  optimization: Distributed machine learning for on-device intelligence, arXiv
  preprint arXiv:1610.02527 (2016).

\bibitem{9533927}
H.~Ren, D.~Anicic, T.~A. Runkler, {{TinyOL: TinyML with Online-Learning on
  Microcontrollers}}, in: 2021 International Joint Conference on Neural
  Networks (IJCNN), 2021, pp. 1--8.
\newblock \href {https://doi.org/10.1109/IJCNN52387.2021.9533927}
  {\path{doi:10.1109/IJCNN52387.2021.9533927}}.

\bibitem{lin2022device}
J.~Lin, L.~Zhu, W.-m. Chen, W.-c. Wang, C.~Gan, S.~Han, On-device training
  under 256kb memory, in: Annual Conference on Neural Information Processing
  Systems, 2022.

\bibitem{lai2018cmsisnn}
L.~Lai, N.~Suda, V.~Chandra, {{CMSIS-NN: Efficient Neural Network Kernels for
  Arm Cortex-M CPUs}} (2018).
\newblock \href {http://arxiv.org/abs/1801.06601} {\path{arXiv:1801.06601}}.

\bibitem{rossi2021vega}
D.~Rossi, F.~Conti, M.~Eggiman, A.~Di~Mauro, G.~Tagliavini, S.~Mach,
  M.~Guermandi, A.~Pullini, I.~Loi, J.~Chen, et~al., {{Vega: A Ten-Core SoC for
  IoT Endnodes With DNN Acceleration and Cognitive Wake-Up From MRAM-Based
  State-Retentive Sleep Mode}}, IEEE Journal of Solid-State Circuits 57~(1)
  (2021) 127--139.

\bibitem{cheng2019distributed}
Z.~Cheng, W.~Wang, Y.~Pan, T.~Lukasiewicz, Distributed low precision training
  without mixed precision (2019).
\newblock \href {http://arxiv.org/abs/1911.07384} {\path{arXiv:1911.07384}}.

\bibitem{kalamkar2019study}
D.~Kalamkar, D.~Mudigere, N.~Mellempudi, D.~Das, K.~Banerjee, S.~Avancha, D.~T.
  Vooturi, N.~Jammalamadaka, J.~Huang, H.~Yuen, J.~Yang, J.~Park, A.~Heinecke,
  E.~Georganas, S.~Srinivasan, A.~Kundu, M.~Smelyanskiy, B.~Kaul, P.~Dubey, A
  study of bfloat16 for deep learning training (2019).
\newblock \href {http://arxiv.org/abs/1905.12322} {\path{arXiv:1905.12322}}.

\bibitem{micikevicius2018mixed}
P.~Micikevicius, S.~Narang, J.~Alben, G.~Diamos, E.~Elsen, D.~Garcia,
  B.~Ginsburg, M.~Houston, O.~Kuchaiev, G.~Venkatesh, H.~Wu, Mixed precision
  training (2018).
\newblock \href {http://arxiv.org/abs/1710.03740} {\path{arXiv:1710.03740}}.

\bibitem{pellegrini2020latent}
L.~Pellegrini, G.~Graffieti, V.~Lomonaco, D.~Maltoni, Latent replay for
  real-time continual learning, in: 2020 IEEE/RSJ International Conference on
  Intelligent Robots and Systems (IROS), IEEE, pp. 10203--10209.

\bibitem{9903209}
F.~De~Vita, G.~Nocera, D.~Bruneo, V.~Tomaselli, M.~Falchetto, On-device
  training of deep learning models on edge microcontrollers, in: 2022 IEEE
  International Conferences on Internet of Things (iThings) and IEEE Green
  Computing \& Communications (GreenCom) and IEEE Cyber, Physical \& Social
  Computing (CPSCom) and IEEE Smart Data (SmartData) and IEEE Congress on
  Cybermatics (Cybermatics), 2022, pp. 62--69.
\newblock \href
  {https://doi.org/10.1109/iThings-GreenCom-CPSCom-SmartData-Cybermatics55523.2022.00018}
  {\path{doi:10.1109/iThings-GreenCom-CPSCom-SmartData-Cybermatics55523.2022.00018}}.

\bibitem{https://doi.org/10.48550/arxiv.2007.11622}
H.~Cai, C.~Gan, L.~Zhu, S.~Han, {{TinyTL: Reduce Activations, Not Trainable
  Parameters for Efficient On-Device Learning}} (2020).
\newblock \href {https://doi.org/10.48550/ARXIV.2007.11622}
  {\path{doi:10.48550/ARXIV.2007.11622}}.

\bibitem{9580920}
L.~Ravaglia, M.~Rusci, D.~Nadalini, A.~Capotondi, F.~Conti, L.~Benini, A tinyml
  platform for on-device continual learning with quantized latent replays, IEEE
  Journal on Emerging and Selected Topics in Circuits and Systems 11~(4) (2021)
  789--802.
\newblock \href {https://doi.org/10.1109/JETCAS.2021.3121554}
  {\path{doi:10.1109/JETCAS.2021.3121554}}.

\bibitem{9604425}
B.~Sudharsan, P.~Yadav, J.~G. Breslin, M.~Intizar~Ali, Train++: An incremental
  ml model training algorithm to create self-learning iot devices, in: 2021
  IEEE SmartWorld, Ubiquitous Intelligence \& Computing, Advanced \& Trusted
  Computing, Scalable Computing \& Communications, Internet of People and Smart
  City Innovation (SmartWorld/SCALCOM/UIC/ATC/IOP/SCI), 2021, pp. 97--106.
\newblock \href {https://doi.org/10.1109/SWC50871.2021.00023}
  {\path{doi:10.1109/SWC50871.2021.00023}}.

\bibitem{https://doi.org/10.48550/arxiv.2201.02863}
J.~Song, F.~Lin, \href{https://arxiv.org/abs/2201.02863}{Pocketnn: Integer-only
  training and inference of neural networks via direct feedback alignment and
  pocket activations in pure c++} (2022).
\newblock \href {https://doi.org/10.48550/ARXIV.2201.02863}
  {\path{doi:10.48550/ARXIV.2201.02863}}.
\newline\urlprefix\url{https://arxiv.org/abs/2201.02863}

\bibitem{https://doi.org/10.48550/arxiv.2206.15472}
J.~Lin, L.~Zhu, W.-M. Chen, W.-C. Wang, C.~Gan, S.~Han, {{On-Device Training
  Under 256KB Memory}} (2022).
\newblock \href {https://doi.org/10.48550/ARXIV.2206.15472}
  {\path{doi:10.48550/ARXIV.2206.15472}}.

\bibitem{9869990}
C.~Cioflan, L.~Cavigelli, M.~Rusci, M.~De~Prado, L.~Benini, Towards on-device
  domain adaptation for noise-robust keyword spotting, in: 2022 IEEE 4th
  International Conference on Artificial Intelligence Circuits and Systems
  (AICAS), 2022, pp. 82--85.
\newblock \href {https://doi.org/10.1109/AICAS54282.2022.9869990}
  {\path{doi:10.1109/AICAS54282.2022.9869990}}.

\bibitem{9797171}
N.~L. Giménez, F.~Freitag, J.~Lee, H.~Vandierendonck, Comparison of two
  microcontroller boards for on-device model training in a keyword spotting
  task, in: 2022 11th Mediterranean Conference on Embedded Computing (MECO),
  2022, pp. 1--4.
\newblock \href {https://doi.org/10.1109/MECO55406.2022.9797171}
  {\path{doi:10.1109/MECO55406.2022.9797171}}.

\bibitem{electronics11040573}
N.~Llisterri~Giménez, M.~Monfort~Grau, R.~Pueyo~Centelles, F.~Freitag,
  On-device training of machine learning models on microcontrollers with
  federated learning, Electronics 11~(4) (2022).
\newblock \href {https://doi.org/10.3390/electronics11040573}
  {\path{doi:10.3390/electronics11040573}}.

\bibitem{AIfES}
\href{https://github.com/Fraunhofer-IMS/AIfES for Arduino}{Aifes for arduino}.
\newline\urlprefix\url{https://github.com/Fraunhofer-IMS/AIfES for Arduino}

\bibitem{10.1007/978-3-031-15074-6_13}
D.~Nadalini, M.~Rusci, G.~Tagliavini, L.~Ravaglia, L.~Benini, F.~Conti,
  {{PULP-TrainLib: Enabling On-Device Training for RISC-V Multi-Core MCUs
  Through Performance-Driven Autotuning}}, in: Embedded Computer Systems:
  Architectures, Modeling, and Simulation: 22nd International Conference, SAMOS
  2022, Samos, Greece, July 3–7, 2022, Proceedings, Springer-Verlag, Berlin,
  Heidelberg, 2022, p. 200–216.
\newblock \href {https://doi.org/10.1007/978-3-031-15074-6_13}
  {\path{doi:10.1007/978-3-031-15074-6_13}}.

\bibitem{NIPS2016_d490d7b4}
A.~N\o~kland,
  \href{https://proceedings.neurips.cc/paper/2016/file/d490d7b4576290fa60eb31b5fc917ad1-Paper.pdf}{Direct
  feedback alignment provides learning in deep neural networks}, in: D.~Lee,
  M.~Sugiyama, U.~Luxburg, I.~Guyon, R.~Garnett (Eds.), Advances in Neural
  Information Processing Systems, Vol.~29, Curran Associates, Inc., 2016.
\newline\urlprefix\url{https://proceedings.neurips.cc/paper/2016/file/d490d7b4576290fa60eb31b5fc917ad1-Paper.pdf}

\bibitem{gholami2021survey}
A.~Gholami, S.~Kim, Z.~Dong, Z.~Yao, M.~W. Mahoney, K.~Keutzer, A survey of
  quantization methods for efficient neural network inference, arXiv preprint
  arXiv:2103.13630 (2021).

\bibitem{LI2020106854}
L.~Li, Y.~Fan, M.~Tse, K.-Y. Lin, A review of applications in federated
  learning, Computers \& Industrial Engineering 149 (2020) 106854.
\newblock \href {https://doi.org/https://doi.org/10.1016/j.cie.2020.106854}
  {\path{doi:https://doi.org/10.1016/j.cie.2020.106854}}.

\bibitem{9060868}
W.~Y.~B. Lim, N.~C. Luong, D.~T. Hoang, Y.~Jiao, Y.-C. Liang, Q.~Yang,
  D.~Niyato, C.~Miao, Federated learning in mobile edge networks: A
  comprehensive survey, IEEE Communications Surveys \& Tutorials 22~(3) (2020)
  2031--2063.
\newblock \href {https://doi.org/10.1109/COMST.2020.2986024}
  {\path{doi:10.1109/COMST.2020.2986024}}.

\bibitem{8662302}
J.~Lee, J.~Lee, D.~Han, J.~Lee, G.~Park, H.-J. Yoo, 7.7 lnpu: A 25.3tflops/w
  sparse deep-neural-network learning processor with fine-grained mixed
  precision of fp8-fp16, in: 2019 IEEE International Solid- State Circuits
  Conference - (ISSCC), 2019, pp. 142--144.
\newblock \href {https://doi.org/10.1109/ISSCC.2019.8662302}
  {\path{doi:10.1109/ISSCC.2019.8662302}}.

\bibitem{9506919}
F.~Montagna, S.~Mach, S.~Benatti, A.~Garofalo, G.~Ottavi, L.~Benini, D.~Rossi,
  G.~Tagliavini, A low-power transprecision floating-point cluster for
  efficient near-sensor data analytics, IEEE Transactions on Parallel and
  Distributed Systems 33~(5) (2022) 1038--1053.
\newblock \href {https://doi.org/10.1109/TPDS.2021.3101764}
  {\path{doi:10.1109/TPDS.2021.3101764}}.

\bibitem{8114708}
V.~Sze, Y.-H. Chen, T.-J. Yang, J.~S. Emer, Efficient processing of deep neural
  networks: A tutorial and survey, Proceedings of the IEEE 105~(12) (2017)
  2295--2329.
\newblock \href {https://doi.org/10.1109/JPROC.2017.2761740}
  {\path{doi:10.1109/JPROC.2017.2761740}}.

\bibitem{8035076}
K.~Osawa, A.~Sekiya, H.~Naganuma, R.~Yokota, Accelerating matrix multiplication
  in deep learning by using low-rank approximation, in: 2017 International
  Conference on High Performance Computing \& Simulation (HPCS), 2017, pp.
  186--192.
\newblock \href {https://doi.org/10.1109/HPCS.2017.37}
  {\path{doi:10.1109/HPCS.2017.37}}.

\bibitem{10.1145/369028.369096}
S.~Huss-Lederman, E.~M. Jacobson, A.~Tsao, T.~Turnbull, J.~R. Johnson,
  \href{https://doi.org/10.1145/369028.369096}{Implementation of strassen's
  algorithm for matrix multiplication}, in: Proceedings of the 1996 ACM/IEEE
  Conference on Supercomputing, Supercomputing '96, IEEE Computer Society, USA,
  1996, p. 32–es.
\newblock \href {https://doi.org/10.1145/369028.369096}
  {\path{doi:10.1145/369028.369096}}.
\newline\urlprefix\url{https://doi.org/10.1145/369028.369096}

\bibitem{pmlr-v80-tschannen18a}
M.~Tschannen, A.~Khanna, A.~Anandkumar,
  \href{https://proceedings.mlr.press/v80/tschannen18a.html}{{S}trassen{N}ets:
  Deep learning with a multiplication budget}, in: J.~Dy, A.~Krause (Eds.),
  Proceedings of the 35th International Conference on Machine Learning, Vol.~80
  of Proceedings of Machine Learning Research, PMLR, 2018, pp. 4985--4994.
\newline\urlprefix\url{https://proceedings.mlr.press/v80/tschannen18a.html}

\bibitem{10.1145/3174243.3174258}
D.~J. Moss, S.~Krishnan, E.~Nurvitadhi, P.~Ratuszniak, C.~Johnson, J.~Sim,
  A.~Mishra, D.~Marr, S.~Subhaschandra, P.~H. Leong,
  \href{https://doi.org/10.1145/3174243.3174258}{A customizable matrix
  multiplication framework for the intel harpv2 xeon+fpga platform: A deep
  learning case study}, in: Proceedings of the 2018 ACM/SIGDA International
  Symposium on Field-Programmable Gate Arrays, FPGA '18, Association for
  Computing Machinery, New York, NY, USA, 2018, p. 107–116.
\newblock \href {https://doi.org/10.1145/3174243.3174258}
  {\path{doi:10.1145/3174243.3174258}}.
\newline\urlprefix\url{https://doi.org/10.1145/3174243.3174258}

\bibitem{9407138}
P.~S. Juan, P.~Alonso-Jordá, E.~S. Quintana-Ortí, High performance and energy
  efficient integer matrix multiplication for deep learning, in: 2021 29th
  Euromicro International Conference on Parallel, Distributed and Network-Based
  Processing (PDP), 2021, pp. 122--125.
\newblock \href {https://doi.org/10.1109/PDP52278.2021.00027}
  {\path{doi:10.1109/PDP52278.2021.00027}}.

\bibitem{8965067}
A.~Garofalo, M.~Rusci, F.~Conti, D.~Rossi, L.~Benini, Pulp-nn: A computing
  library for quantized neural network inference at the edge on risc-v based
  parallel ultra low power clusters, in: 2019 26th IEEE International
  Conference on Electronics, Circuits and Systems (ICECS), 2019, pp. 33--36.
\newblock \href {https://doi.org/10.1109/ICECS46596.2019.8965067}
  {\path{doi:10.1109/ICECS46596.2019.8965067}}.

\bibitem{9935273}
D.~Han, S.~Kang, S.~Kim, J.~Lee, H.-J. Yoo, Energy-efficient dnn training
  processors on micro-ai systems, IEEE Open Journal of the Solid-State Circuits
  Society 2 (2022) 259--275.
\newblock \href {https://doi.org/10.1109/OJSSCS.2022.3219034}
  {\path{doi:10.1109/OJSSCS.2022.3219034}}.

\bibitem{9381618}
A.~Burrello, A.~Garofalo, N.~Bruschi, G.~Tagliavini, D.~Rossi, F.~Conti, Dory:
  Automatic end-to-end deployment of real-world dnns on low-cost iot mcus, IEEE
  Transactions on Computers 70~(8) (2021) 1253--1268.
\newblock \href {https://doi.org/10.1109/TC.2021.3066883}
  {\path{doi:10.1109/TC.2021.3066883}}.

\bibitem{https://doi.org/10.48550/arxiv.1704.04861}
A.~G. Howard, M.~Zhu, B.~Chen, D.~Kalenichenko, W.~Wang, T.~Weyand,
  M.~Andreetto, H.~Adam, \href{https://arxiv.org/abs/1704.04861}{Mobilenets:
  Efficient convolutional neural networks for mobile vision applications}
  (2017).
\newblock \href {https://doi.org/10.48550/ARXIV.1704.04861}
  {\path{doi:10.48550/ARXIV.1704.04861}}.
\newline\urlprefix\url{https://arxiv.org/abs/1704.04861}

\end{thebibliography}
\end{document}